\pdfoutput=1

\documentclass[11pt]{article}

\usepackage[final]{acl}

\usepackage{paralist} 
\usepackage{times}
\usepackage{latexsym}
\usepackage{booktabs} 
\usepackage{amssymb}
\usepackage{multirow}
\usepackage{hyperref}
\usepackage{url}
\usepackage{adjustbox}
\usepackage{pgf-pie}
\usepackage{tcolorbox}
\usepackage{listings}
\usepackage{pgfplots}
\usepackage{tikz}
\usetikzlibrary{arrows}
\usepackage{framed}
\usepackage{subcaption}
\usepackage{booktabs}
\usepackage[utf8]{inputenc}
\usepackage{tabularx}
\usepackage{longtable}
\usepackage{wrapfig,lipsum,booktabs}
\usepackage{multicol, multirow}
\usepackage{seqsplit}
\usepackage{xcolor}
\usepackage[table]{xcolor}
\usepackage{tabularray}
\usepackage{amsmath}
\usepackage{cleveref}
\usepackage[T1]{fontenc}

\usepackage[utf8]{inputenc}

\usepackage{microtype}

\usepackage{inconsolata}

\usepackage{graphicx}

\definecolor{rosso}{RGB}{220,57,18}
\definecolor{giallo}{RGB}{255,153,0}
\definecolor{blu}{RGB}{102,140,217}
\definecolor{verde}{RGB}{16,150,24}
\definecolor{viola}{RGB}{153,0,153}
\definecolor{yellow}{RGB}{255,255,168}
\definecolor{lightblue}{RGB}{0, 255, 255}
\definecolor{green2}{RGB}{64, 224, 208}
\definecolor{orange}{RGB}{255, 195, 0}
\definecolor{grey}{RGB}{220,220,220}
\definecolor{lightgreen}{RGB}{144,238,144}
\definecolor{pink}{RGB}{255, 192, 203}
\definecolor{red}{RGB}{255,71,77}

\usepackage{tcolorbox}

\definecolor{myboxcolor}{RGB}{254,254,254} 
\definecolor{myframe}{RGB}{0,0,128} 

\newtcolorbox{mybanner}{
  colback=myboxcolor,
  colframe=myframe,
  boxrule=1pt, 
  left=1pt,
  right=1pt,
  top=1pt,
  bottom=1pt,
}

\newtcolorbox{mybody}{
  colback=myboxcolor,
  colframe=myframe,
  boxrule=1pt, 
  left=1pt,
  right=1pt,
  top=1pt,
  bottom=1pt,
}

\definecolor{DarkGreen}{RGB}{30,130,30}
\newcommand{\cmark}{\textcolor{DarkGreen}{\ding{51}}}
\newcommand{\xmark}{\textcolor{red}{\ding{55}}}%
\newcommand{\mapagent}{{MapAgent}~}
\newcommand{\mapagentnospace}{{MapAgent}}

\newcommand{\toolnospace}{{MapEval}}
\newcommand{\visualtool}{{MapEval-Visual}~}

\newcommand{\textualtool}{{MapEval-Textual}~}

\newcommand{\apieval}{{MapEval-API}~}
\newcommand{\apitool}{{MapEval-API}~}

\usepackage{pifont}

\title{\mapagentnospace: A Hierarchical Agent for Geospatial Reasoning with Dynamic Map Tool Integration}

\author{
    Md Hasebul Hasan$^{1,2}$\quad
    Mahir Labib Dihan$^{1}$ \quad
    \textbf{Tanzima Hashem}$^{1}$ \quad
    \\
    \textbf{Mohammed Eunus Ali}$^{3}$ \quad 
    \textbf{Md Rizwan Parvez}$^{4}$ \quad \\
    $^1$Bangladesh University of Engineering and Technology (BUET) \\
    $^2$University of Texas at Arlington\quad 
    $^3$Faculty of Information Technology, Monash University \quad \\
    $^4$Qatar Computing Research Institute (QCRI) \\
    { \{hasebulhasan97, mahirlabibdihan\}@gmail.com},  \\
    {tanzimahashem@cse.buet.ac.bd, eunus.ali@monash.edu, mparvez@hbku.edu.qa}
}

\begin{document}

\maketitle

\noindent Agentic AI has significantly extended the capabilities of large language models (LLMs) by enabling complex reasoning and tool use. However, most existing frameworks are tailored to domains such as mathematics, coding, or web automation, and fall short on geospatial tasks that require spatial reasoning, multi-hop planning, and real-time map interaction. 
To address these challenges, we introduce {\bf \mapagentnospace}, a hierarchical multi-agent plug-and-play framework with customized toolsets and agentic scaffolds for map-integrated geospatial reasoning. Unlike existing flat agent-based approaches that treat tools uniformly—often overwhelming the LLM when handling similar but subtly different geospatial APIs—\mapagent decouples planning from execution. A high-level planner decomposes complex queries into subgoals, which are routed to specialized modules. For tool-heavy modules—such as map-based services—we then design a dedicated map-tool agent that efficiently orchestrates related APIs adaptively in parallel to effectively fetch geospatial data relevant for the query, while simpler modules (e.g., solution generation or answer extraction) operate without additional agent overhead. This hierarchical design reduces cognitive load, improves tool selection accuracy, and enables precise coordination across similar APIs. We evaluate \mapagent on four diverse geospatial benchmarks—MapEval-Textual, MapEval-API, MapEval-Visual, and MapQA—and demonstrate substantial gains over state-of-the-art tool-augmented and agentic baselines. We open-source our framwork at https://github.com/Hasebul/MapAgent.
\section{Introduction}
\label{sec:introduction}
\begin{figure}[h!]
    \centering
    \includegraphics[width=1\linewidth]{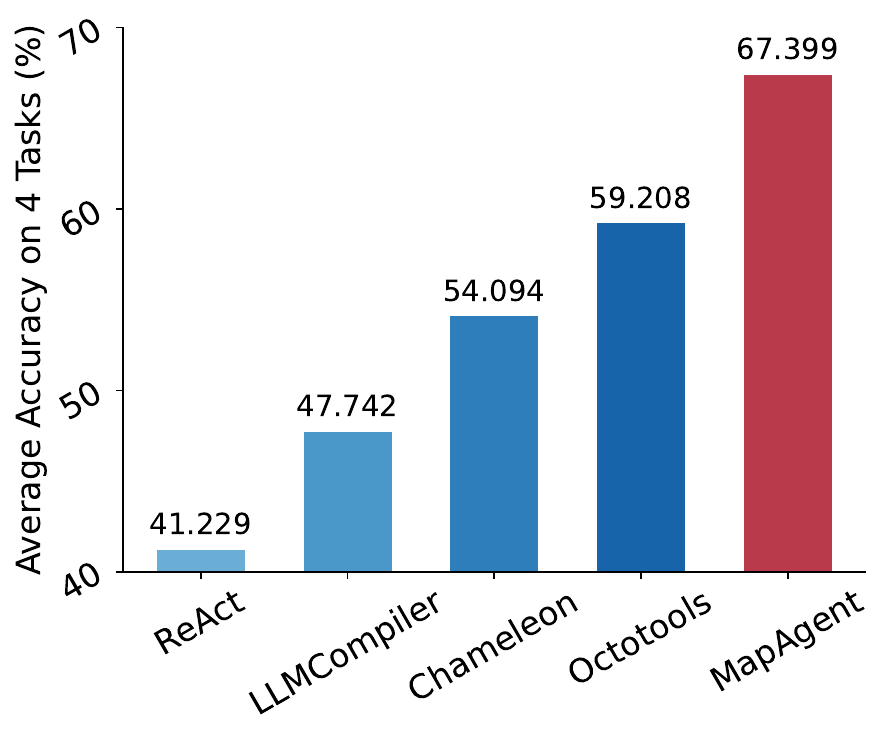}
    \caption{Performance comparison across 4 geo-spatial benchmarks.  \mapagent  significantly outperforms others and achieves state-of-the-art (SOTA) performance. }
    \label{fig:overview}
\vspace{-10pt}
\end{figure}
\begin{figure*}[h]
    \centering
     \includegraphics[width=1.0\linewidth]{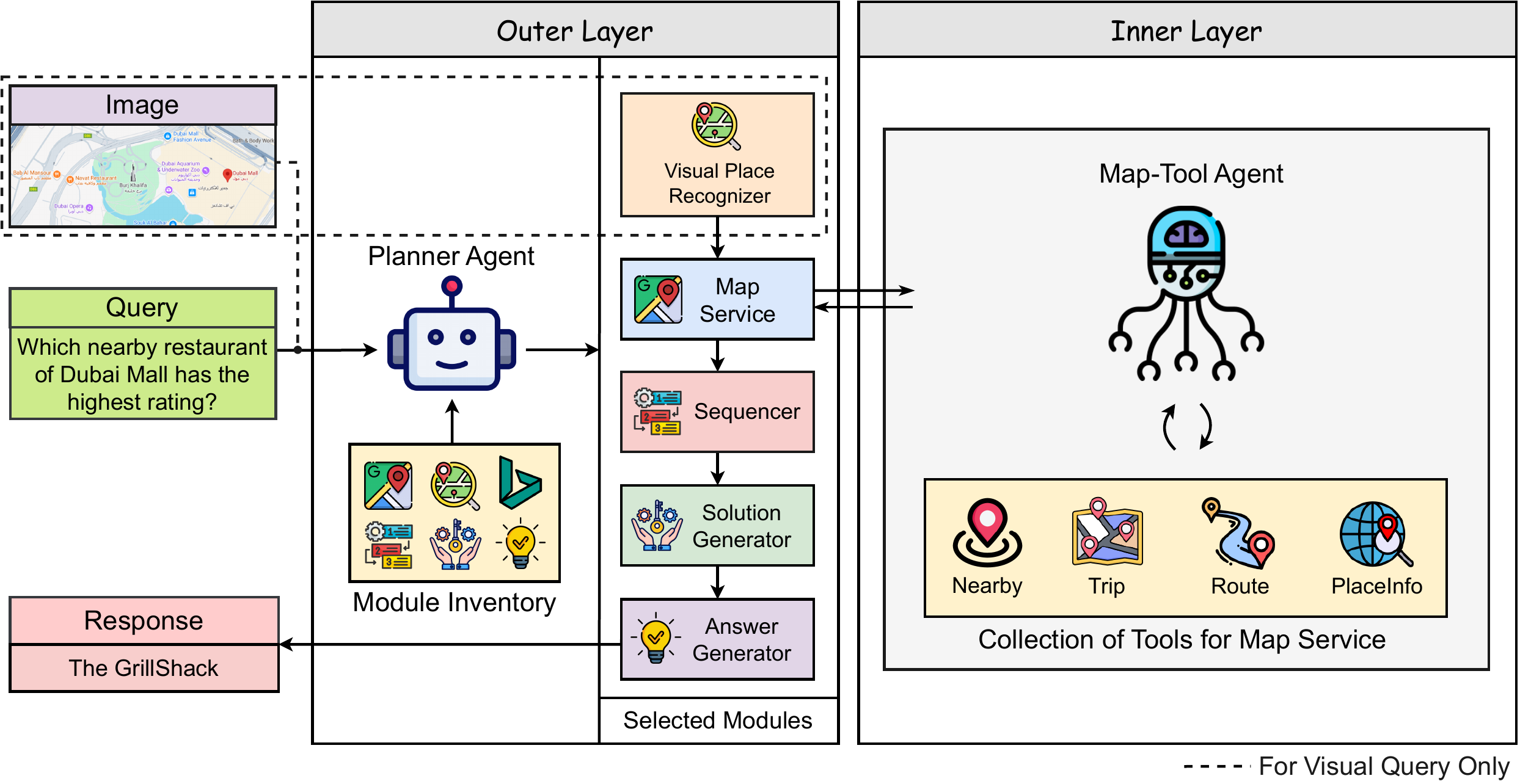}
     \vspace{-10pt}
    \caption{{Overview of \mapagentnospace.} Given a user query (optionally with an image), the Planner Agent decomposes it into subtasks using the available Module Inventory and selects appropriate modules for each subtask. For tool-heavy modules, a dedicated agent (e.g., a Map-Tool Agent) manage interactions with the associated tools adaptively.  
}
    \vspace{-10pt}
    \label{fig:overview-fig2}
\end{figure*}
Agentic large language models (LLMs) have significantly expanded the scope of AI systems by enabling complex reasoning, subgoal decomposition, and dynamic tool use~\cite{yao2023react,kim2024llm,du2024anytool}. Despite these advances, most existing frameworks predominantly focus on domains such as mathematics~\cite{lu2023chameleon,lu2025octotools}, software engineering~\cite{jimenez2023swe}, and web automation~\cite{qin2023toolllm,du2024anytool}, while their application to \textbf{geospatial reasoning}—a ubiquitous capability that enables the automation of everyday tasks through natural language instructions, particularly in scenarios requiring interaction with specialized tools like map services—remains limited. This includes tasks such as route planning, location-based decision-making, spatial comparison, and finding nearby points of interest (e.g., restaurants, gas stations, or EV charging points) that satisfy user constraints.

Geospatial tasks pose distinct challenges compared to other reasoning problems. Queries in this domain often require multi-hop planning, dynamic spatial grounding, and coordination across multiple external APIs with overlapping but subtly different capabilities. For instance, solving queries like \textit{“Find the shortest route from home to the office with a stop at a highly rated coffee shop”} or \textit{“Plan a three-day road trip from San Francisco to Yosemite with overnight stops at scenic towns and hiking spots along the way”} involves chaining services for nearby place search, distance estimation, POI (point-of-interest) retrieval, temporal scheduling, and real-time filtering. Despite addressing very related geospatial functions, the available Map APIs vary in their own schema, constraints, and precious functionalities, making it non-trivial to integrate them effectively. The agent must not only orchestrate multiple tools, but also combine and reason over the retrieved data to produce timely, coherent responses that preserve geospatial consistency.
\begin{table*}[t]
    \centering
    \begin{tabular}{@{}l|c|c|c|c@{}}
        \hline
        
        \multirow{2}{*}{\textbf{Method}} & {\textbf{Parallel}} & \textbf{Compositional} & \multirow{2}{*}{\textbf{Hierarchical}} &  {\textbf{Plug-n-play}} \\ 
         & \textbf{Tool Calling} & \textbf{Reasoning} &  & \textbf{or Map-tools} \\
        \hline
        ReAct \cite{yao2023react}        & \xmark & \xmark & \xmark & \xmark \\
        LLMCompiler \cite{kim2024llm}    & \cmark & \xmark & \xmark & \xmark \\
        Octotools \cite{lu2025octotools} & \cmark & \cmark & \xmark & \cmark \\
        Chameleon \cite{lu2023chameleon} & \xmark & \cmark & \xmark & \cmark \\
        Anytool \cite{du2024anytool}     & \xmark & \xmark & \cmark & \xmark \\
        \textbf{\textit{MapAgent (Ours)}}       & \cmark & \cmark & \cmark & \cmark \\
        \hline
    \end{tabular}
    \caption{Feature comparison of different tool calling frameworks in context to map-integrated geospatial reasoning.}
    \label{tab:framework_comparison}
\end{table*}
However, current agentic systems are not designed for this setting. Most plug-and-play tool-using agents adopt flat execution architectures that treat tools as generic, interchangeable black boxes. This leads to two major limitations in the context of geospatial reasoning: 
\emph{(i) Tool inflation:} Existing systems often bundle disparate tools—one of each kind—without accounting for the fine-grained functional variations present within map services. As geospatial APIs proliferate (e.g., distance, proximity, directions, routing, nearby search, place details, timelines), agents face a combinatorial burden in both planning and execution, increasing decision complexity and reducing overall effectiveness. 

\emph{(ii) Tool incapability:} Tools integrated into existing agents are typically primitive (single-API-based) and generic (e.g., image captioning, table lookup, code generation, or web search), and are not designed for the rich, mixed-mode (parallel and sequential) interactions required by real-world map services like Google Maps. While some API calls—such as computing distances and retrieving place details—can be performed in parallel, many tasks require tightly coordinated sequential steps, such as fetching detailed information only for places identified in a prior nearby search.

To address these challenges, we propose \textbf{\mapagentnospace}, a hierarchical plug-and-play multi-agent framework designed for map-integrated geospatial reasoning.
First, to overcome the issue of \emph{tool incapability}, we design a set of four heterogeneous map tools, each composed of one or more primitive APIs tailored to perform key geospatial operations. These tools encapsulate common spatial functionalities—such as nearby search, route planning, and place detail retrieval—and are constructed to support both parallel and sequential execution flows. By abstracting low-level primitive API calls into higher-level tool interfaces, \mapagent{} enables robust handling of mixed-mode reasoning: for example, retrieving details of candidate places found via a nearby search in a sequential pipeline, while simultaneously computing alternative routes in parallel. 
This structured orchestration supports the complex, map-centric reasoning patterns required for real-world spatial tasks.

Second, to mitigate the problem of \emph{tool inflation}, \mapagent{} adopts a hierarchical architecture that decouples high-level task planning from low-level tool execution. A top-level planner agent decomposes complex natural language queries into structured subgoals and routes them to appropriate tool workflows. For tool-intensive subgoals—such as those involving map services—we introduce a dedicated \textit{map-tool agent} that adaptively manages interactions with multiple map tools, issuing parallel API calls where appropriate and coordinating sequential workflows when required. In contrast, lightweight tasks—such as answer formatting or preference synthesis—are handled directly without additional agent overhead.\\
\indent We evaluate \mapagent{} across four diverse geospatial reasoning benchmarks—MapQA, MapEval-Textual, MapEval-Visual, and MapEval-API—each presenting distinct challenges, including textual context, multimodal inputs, and query-only settings. Built using both open-weight and closed-source backbone LLMs, \mapagent{} consistently outperforms prior state-of-the-art agentic and tool-augmented LLM frameworks across all benchmarks. Utilizing GPT-3.5-Turbo, it achieves a 10\% improvement on both the MapEval-API and MapEval-Textual datasets, and an 11.22\% improvement on the MapQA dataset over the strong baseline OctoTools~\cite{lu2025octotools}. Furthermore, with GPT-4o, \mapagent{} achieves a 4.41\% improvement on the MapEval-Visual dataset compared with the same baseline. Overall, \mapagent{} improves performance by 8.2\% on average over the strongest baseline, OctoTools, in geospatial tasks that require multi-hop planning, spatial inference, and dynamic tool use. Our ablations, qualitative, and fine-grained analyses further provide more insights into the effectiveness of \mapagent{}.

\section{Related Work}
\paragraph{Compositional Reasoning with LLMs.}
Solving complex multi-step problems often involves decomposing tasks into manageable sub-problems \cite{perez2020unsupervised,khot2023decomposed}. Prompt-based methods like Chain-of-Thought \cite{wei2022chain}, Least-to-Most \cite{zhou2023leasttomost}, ReAct \cite{yao2023react}, Pearl \cite{sun2024pearl}, Forest-of-Thought \cite{bi2024forest}, and rStar-Math \cite{guan2025rstar} have advanced the ability of LLMs to reason sequentially. Other approaches synthesize structured programs from natural language (e.g., LLM Compiler \cite{kim2024llm}) or orchestrate modular tools (e.g., Chameleon \cite{lu2023chameleon}, OctoTools \cite{lu2025octotools}). While effective, these systems often assume flat or static module inventories and struggle with tool variants that differ subtly in schema or functionality. In contrast, \mapagent leverages hierarchical planning and dynamic module selection to support plug-and-play coordination over overlapping but heterogeneous APIs, particularly in the geospatial domain.

\paragraph{Geospatial Reasoning with Language Models.}
Geospatial QA has been studied through rule-based and template-driven systems that convert natural language into structured queries over static databases (e.g., PostGIS, DBpedia \cite{auer2007dbpedia}, YAGO \cite{suchanek2007yago}, Freebase \cite{bollacker2007freebase}, OSM). Notable examples include GeoSPARQL-based systems \cite{car2022ogc} and datasets such as GeoQuestions1089 \cite{kefalidis2023benchmarking} and YAGO2geo \cite{karalis2019extending}. These methods are precise but inflexible, limited by their reliance on static schemas and query grammars.

Recent work explores LLMs' intrinsic geographic knowledge (e.g., GPT4GEO \cite{roberts2023gptgeo}) or their capacity to answer map-related queries directly (e.g., MapQA \cite{li2025mapqa}, MapEval \cite{mahir2025mapeval, dihan-etal-2025-mapqator}). While promising, these models lack mechanisms to integrate external tools or coordinate multi-API workflows. \mapagent builds on this line of work by introducing a multi-agent, tool-augmented framework that dynamically coordinates geospatial modules across modalities—including textual, visual, and API inputs—offering enhanced reasoning capabilities for real-world map-based tasks.

\section{Method}
\label{sec:method}
We propose \textbf{\mapagentnospace}, a hierarchical plug-and-play multi-agent framework for map-integrated geospatial reasoning. \mapagent{} addresses two core limitations in current agentic systems—\emph{tool incapability} and \emph{tool inflation}—through a structured two-layer scaffold that cleanly separates reasoning, planning, and tool execution.

\subsection{Overview}
\label{sec:method:overview}

We introduce \textbf{\mapagentnospace}, a hierarchical multi-agent framework for solving complex geospatial reasoning tasks via real-world map services. Real-world queries often require chaining spatial operations, querying external APIs, and integrating geographic evidence—capabilities that single-shot prompting lacks. To address this, \mapagent{} enables structured planning, map-tool composition, and spatial reasoning through specialized agents orchestrated in a hierarchical coordination.

In \mapagentnospace, a top-level \textit{planner agent} $\mathcal{P}$ first decomposes the user query $x$ into subgoals, enabling modular execution and interpretable reasoning. These are handled by specialized \textit{functional modules} $M$,  which delegate tasks such as filtering or aggregation to dedicated components. \Cref{fig:overview} shows an overview of \mapagentnospace.

Geospatial subgoals are routed to a \textit{Map-Tool Agent} $\mathcal{M}_{\text{map}}$, which composes four core tools \texttt{Nearby}, \texttt{PlaceInfo}, \texttt{Route}, and \texttt{Trip}—each built over real-world Google Map APIs to support spatial search, detail retrieval, and routing. This modular design allows \mapagent to solve complex, multi-hop spatial queries by dynamically planning over tool-based workflows. We describe the architecture in \Cref{sec:architecture}, the planner agents and modules in \Cref{subsec:modules}, the map tools in \Cref{subsec:maptools}.
\begin{table*}[t]
    \centering
    \includegraphics[width=1\linewidth]{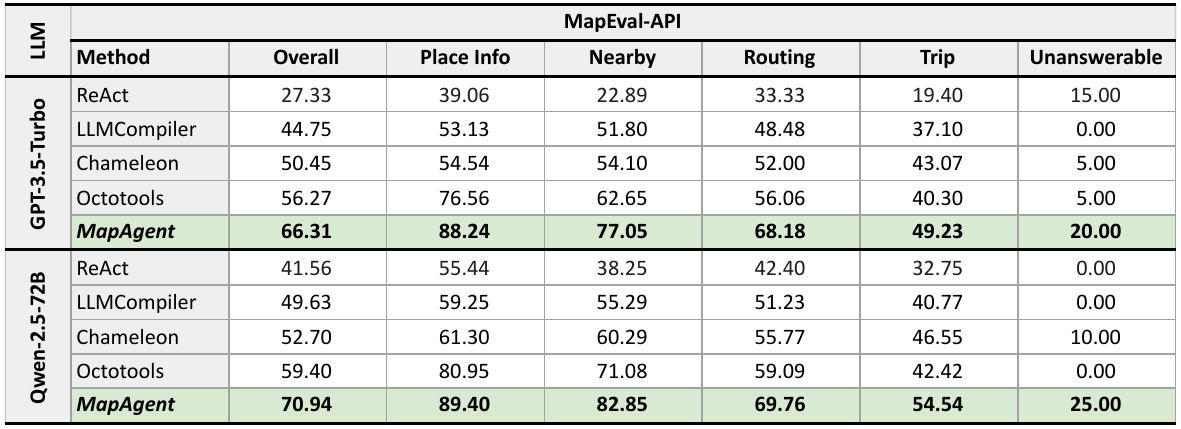}
    \caption{Accuracy (\%) of different methods and backbones on the MapEval-API Dataset.}
    \label{fig:mapeval-api-result}
\end{table*}
\begin{table*}[t]
    \centering
    \includegraphics[width=1\linewidth]{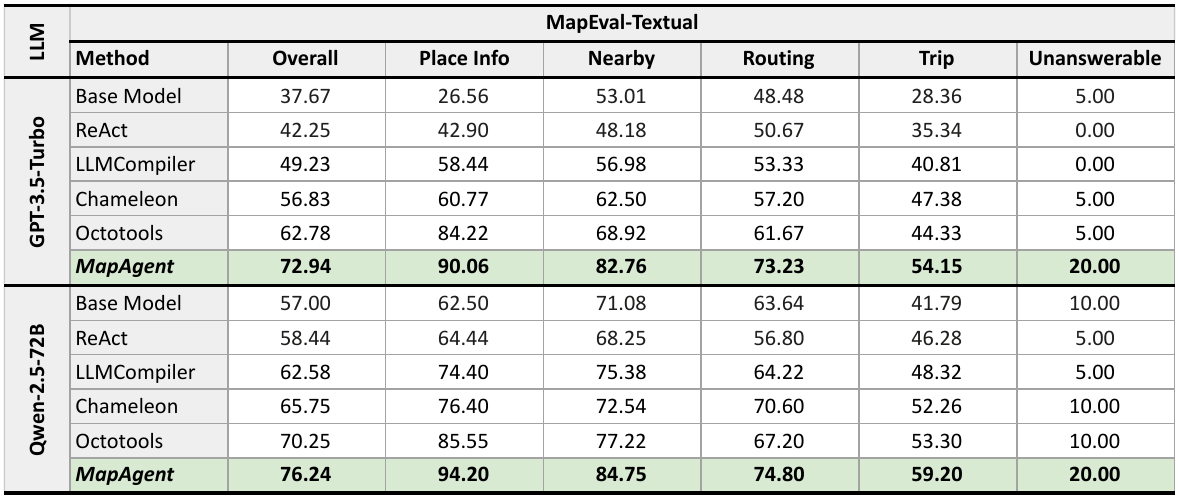}
    \caption{Accuracy (\%) of different methods and backbones on the MapEval-Textual Dataset.}
    \label{fig:mapeval-textual-result}
    \vspace{-10pt}
\end{table*}

\subsection{Scaffolded Architecture (Hierocracy)}
\label{sec:architecture}
\mapagent{} comprises two conceptual layers:
\paragraph{Planner Layer (Top-level):} A planner agent decomposes the input $x$ into a sequence of semantically coherent subgoals $[g_1, \dots, g_n]$, and routes each $g_i$ to a functional module or tool chain based on a structured inventory (discussed in \Cref{tab:modules}).
\paragraph{Execution Layer (Bottom-level):} This layer manages module-specific execution logic. It includes both lightweight modules (e.g., sequencing, formatting) and a specialized \textit{Map-Service Module}, which integrates a dedicated \textbf{Map-Tool Agent} responsible for executing geospatial tasks using a curated set of map tools.

\subsection{Module Inventory and Planner Agent}
\label{subsec:modules}
The planner agent $\mathcal{P}$ (based on an LLM or VLLM) receives the input $x$, an inventory $M$ of available module to perform the subgoals, and  outputs a structured execution plan $\pi = [(g_1, m_1), ..., (g_n, m_n)]$, where each $g_i$ is a subgoal assigned to a module $m_i \in M$. {Module Inventory} $M$ contains several task-specific modules:

\noindent
 \textbf{Visual Place Recognizer}:  Given an image map-snapshot, using the corresponding VLLM, this module extracts the central geographic place (name) in the image and estimates the geographic scope (approximate radius) covered in the image. 
\noindent \textbf{Sequencer:} This module is responsible for organizing and structuring the responses received from preceding modules.  It arranges unstructured information into a logical sequence using LLMs. 

\noindent \textbf{Solution Generator:} This module synthesizes the final answer to the user query using the corresponding LLM or VLM. It takes structured information and generates a comprehensive response. 

\noindent \textbf{Answer Generator:} This module focuses on refining and verifying the answer generated by the Solution Generator. It extracts the predicted answer and checks its consistency.

\noindent  \textbf{Map-Service Module}: A tool-heavy module managed by the Map-Tool Agent to handle core Map-tool interactions.

\subsection{Map-Service Module: Map-Tool Agent and Tool Design}
\label{subsec:maptools}
Map-Tool Agent $\mathcal{M}_{\textbf{map}}$ takes as input the query $x$, along with an optional place name and radius inferred by the Visual Place Recognizer for multimodal queries. It then adaptively interacts with a suite of geospatial tools designed to fetch the necessary spatio-temporal data from underlying map services (e.g., Google Maps).

We design four heterogeneous \textbf{map tools}, each composed of one or more primitive APIs from Google Map (e.g., directions, place info) to support essential geospatial operations:\\
\textbf{Trip Tool}: Retrieves information about locations and travel routes between them. Combines the Place Details and Directions APIs to fetch place metadata and step-by-step route instructions. \\
\textbf{Route Tool}: Fetches route information between two points using the Directions API, including distance, estimated time, and navigation steps. \\
\textbf{Nearby Tool}: Identifies points of interest within a specified area using the Nearby Search API. Returns place names, ratings, and other metadata. \\
\textbf{PlaceInfo Tool}: Retrieves detailed information about a given place using the Place Details API, including address, hours, contact info, and reviews. \\

\section{Experimental Setup}

\subsection{Datasets}
We evaluated \mapagent on four benchmarks spanning two modalities as shown in Table \ref{tab:dataset_summary}.  These benchmarks encompass a wide range of complex geospatial reasoning tasks, including
long-context reasoning, API interactions, visual map analysis, numerical calculation, and multi-step reasoning. Three benchmarks are from MapEval \cite{mahir2025mapeval}:
\begin{table}[h!]
\centering
\small
\begin{tabular}{l|c|l}
\hline
\textbf{Dataset} & \textbf{\# Tasks} & \textbf{Modality} \\
\hline
MapEval-Textual & 300  & Text \\
MapEval-API  & 300 & Text \\
MapEval-Visual & 400  &  Vision \\
MapQA & 3154 & Text \\
\hline
\end{tabular}
\caption{Overview of datasets: statistics and modality}
\label{tab:dataset_summary}
\vspace{-10pt}
\end{table}

    \noindent\textbf{MapEval-Textual:} Given a user query with a long textual context describing map locations, POIs, routes, navigation details, travel distances/times, and user-generated content such as ratings/reviews, the task is to answer the query based on the context. \\

\begin{table*}[t]
   
    \centering
    \includegraphics[width=1\linewidth]{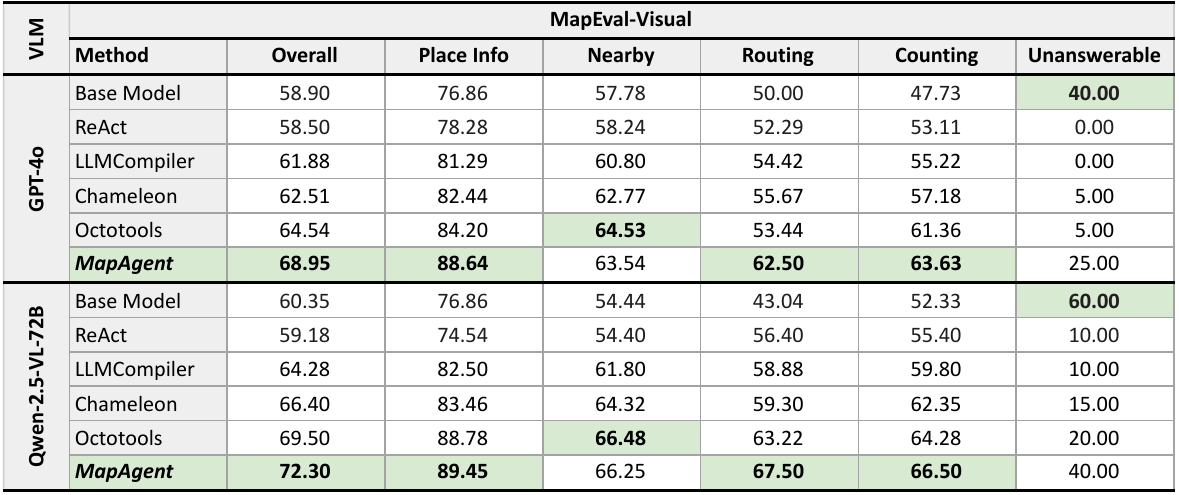}
    \caption{Accuracy (\%) of different methods and backbones on the MapEval-Textual Dataset.}
    \label{fig:mapeval-visual-result}
    \vspace{-10pt}
\end{table*}
    \noindent\textbf{MapEval-API:} Given a user query and access to map-based APIs, the task is to answer the query based on the structured data retrieved via the APIs.\\
    \textbf{MapEval-Visual:} Given a user query and a digital map snapshot showing spatial layouts, routes, landmarks, OCR text (e.g., ratings), and symbolic elements (e.g., icons or signs), the task is to answer the query based on the visual context. \\
    \textbf{MapQA~\cite{li2025mapqa}:} A geospatial QA benchmark constructed from OpenStreetMap using SQL query templates. It includes QA pairs covering nine reasoning types (e.g., neighborhood inference, spatial proximity, type classification), grounded in real-world geo-entity geometries.

\begin{table*}[t]
    \centering
    \includegraphics[width=1\linewidth]{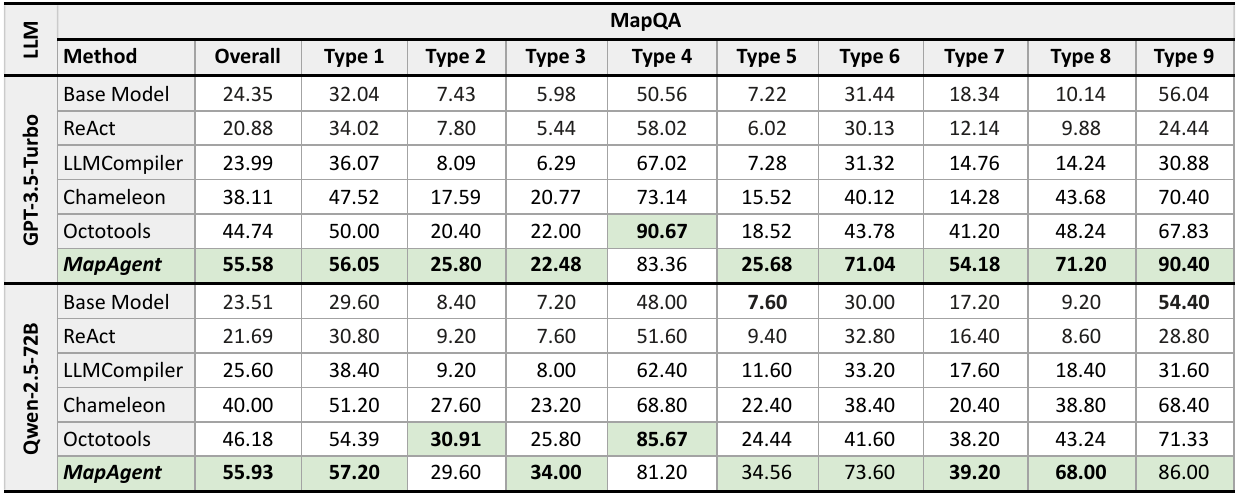}
    \caption{Accuracy (\%) of different methods and backbones on the MapQA Dataset.}
    \label{fig:map-qa-result}
\end{table*}

\begin{table*}[h!]
\centering
\begin{tabular}{|c|c|c|c|c|}
\hline
\textbf{Preemptive Geospatial Tools} & \textbf{Our Maptools} & \textbf{Hierarchical} & \textbf{Accuracy (\%)} & \textbf{Performance } \\
\textbf{(Google Map APIs)} & & \textbf{Architecture} & & \textbf{Drop (\%)} \\
\hline
\cmark & \xmark & \xmark & 39.37 & 49.00 \\
\hline
\cmark & \cmark & \xmark & 56.39 & 27.18 \\
\hline
\cmark & \cmark & \cmark & 77.44 & - \\
\hline
\end{tabular}
\caption{Query-solving accuracy with and without the proposed Maptools and Hierarchical Architecture}
\label{tab:ablation}
\end{table*}

\subsection{Baseline and Metric}
\mapagent is a multi-agent compositional reasoning framework. We therefore compare its performance against agentic frameworks, such as ReAct \cite{yao2023react} and LLMCompiler \cite{kim2024llm}, and compositional reasoning frameworks, such as Chameleon \cite{lu2023chameleon} and OctoTool \cite{lu2025octotools}. For the \apieval, \textualtool and MapQA datasets, we use GPT-3.5-Turbo \citep{openai2022chatgpt} and Qwen-2.5-72b \cite{qwen2.5}. For the \visualtool dataset, we use models served via vLLM, specifically GPT-4o \cite{gpt4o} and Qwen-2.5-VL-72b \citep{bai2025qwen2}. We evaluate model performance using accuracy, expressed as a percentage.

\subsection{Results}

 The evaluation of the \mapagent framework across four distinct and challenging map-related datasets reveals its significant advancements in accuracy, consistently outperforming existing agentic approaches and demonstrating robust performance irrespective of the underlying language model backbone. These findings underscore \mapagent's effectiveness as a robust solution for a broad spectrum of map-based query processing tasks.

\paragraph*{\textbf{Strong Gains Across Benchmarks}}
 Across all evaluated datasets, the MapAgent framework consistently achieves high accuracy. On the MapEval-API dataset in Table \ref{fig:mapeval-api-result}, MapAgent achieves overall accuracies of roughly 70\% and 72\% when powered by \texttt{GPT-3.5-Turbo} and \texttt{Qwen-2.5-72B}, respectively. Similarly, on the MapEval-Textual dataset in Table \ref{fig:mapeval-textual-result}, MapAgent attains accuracies of 72.94\% with \texttt{GPT-3.5-Turbo} and 76.24\% with \texttt{Qwen-2.5-72B}. The MapEval-Visual results in Table \ref{fig:mapeval-visual-result} further reinforce this trend, with \mapagent achieving the highest overall accuracy--68.95\% using \texttt{GPT-4o} and 72.30\% with \texttt{Qwen-2.5-VL-72B}. Finally, on the MapQA benchmark in Table \ref{fig:map-qa-result}, MapAgent achieves the highest overall accuracy of 55.58\% with \texttt{GPT-3.5-Turbo} and 55.93\% with \texttt{Qwen-2.5-72B}.

\paragraph*{\textbf{Surpassing Prior Methods}}
The performance of \mapagent consistently surpasses that of other prominent methods, including OctoTools, Chameleon, ReAct, and LLMCompiler---notably the second-best performing OctoTools---across all evaluated datasets and language models. In Figure \ref{fig:mapagent_vs_octotools}, on the MapEval-API dataset, \mapagent shows accuracy gains of approximately 10\% and 11.54\% over OctoTools when using \texttt{GPT-3.5-Turbo} and \texttt{Qwen-2.5-72B}, respectively. For the MapQA benchmark, \mapagent outperforms OctoTools by approximately 11.22\% with \texttt{GPT-3.5-Turbo} and 9\% with \texttt{Qwen-2.5-72B}. Lastly, on the MapEval-Textual dataset, \mapagentnospace's accuracy is nearly 10\% and 6\% higher than OctoTools with \texttt{GPT-3.5-Turbo} and \texttt{Qwen-2.5-72B}. In the MapEval-Visual evaluations, \mapagent outperforms OctoTools by about 4.41\% with \texttt{GPT-4o} and 2.8\% with \texttt{Qwen-2.5-VL-72B}.

\paragraph*{\textbf{Backbone Agnostic}}
A key strength of the MapAgent framework is its consistent high performance across different language model backbones. When evaluated on the MapEval-API and MapEval-Textual datasets, MapAgent demonstrates superior accuracy with both \texttt{GPT-3.5-Turbo} and \texttt{Qwen-2.5-72B} models. This is further evident in the MapEval-Visual dataset, where MapAgent achieves the highest accuracy with both \texttt{GPT-4o} and \texttt{Qwen-2.5-VL-72B} vision-language models. The consistent top-tier results across diverse LLMs and VLMs highlight MapAgent’s robustness and adaptability, showing its effectiveness is not tied to any specific model architecture.
 \section{Ablation and Qualitative Analysis}
\label{appendix ablation}
\subsection{Impact of Our Maptools and Hierarchical Architecture}
To evaluate the impact of our proposed Maptools and Hierarchical Architecture, we conducted an ablation study on query solving accuracy with and without them. The results of this study are summarized in Table \ref{tab:ablation}.

As shown in the table, when both our custom-developed tools and the hierarchical architecture are utilized, the solving accuracy reaches 77.44\%. However, removing the hierarchical architecture causes a performance drop of 27.18\%, lowering the accuracy to 56.39\%. Finally, removing both the hierarchical architecture and our map tools results in an additional performance drop of 22\%, with the final accuracy decreasing to 39.37\%. These results indicate the significance of the hierarchical architecture and our built-in map tools.

\subsection{Why does using visuals with an API perform better than simply passing visual questions to VLLM?} 
\begin{figure}[t]
\vspace{-10pt}
    \centering
    \includegraphics[width=0.9\linewidth]{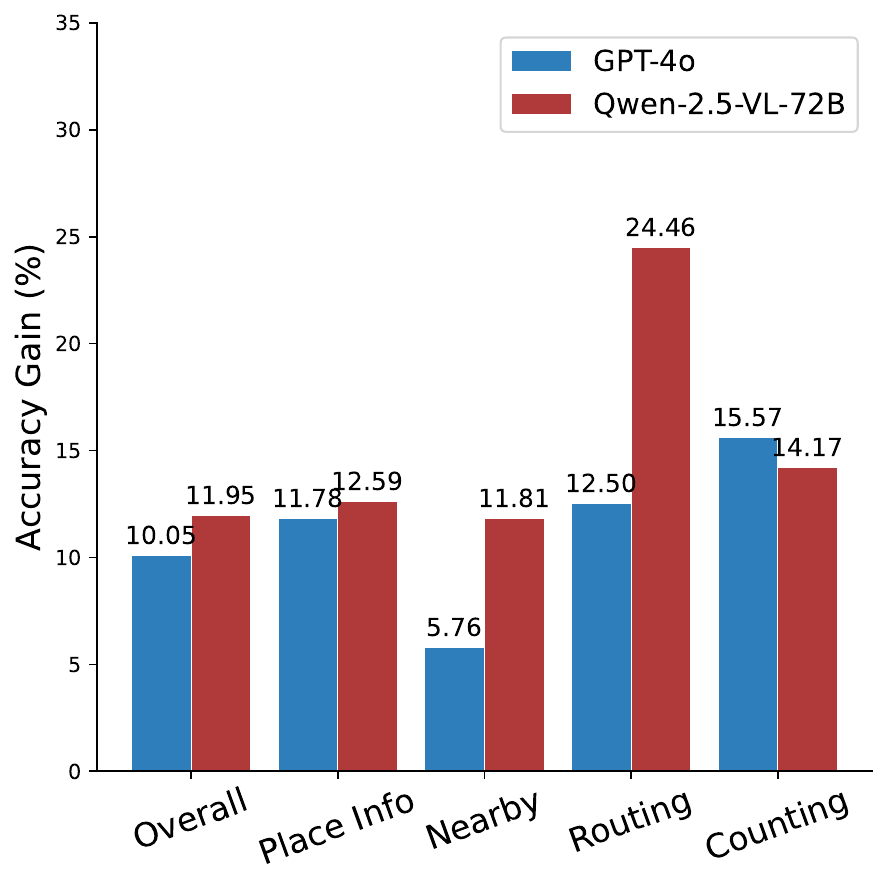}
    \caption{Performance gain from API-based visuals.}
    \label{fig:visual_improve}
\end{figure}

When MapAgent attempts to solve a visual query, it utilizes Visual Place Recognizer module to obtain information about the image's center and estimates its boundaries. Subsequently, the Google Maps module uses this information to fetch relevant data for resolving the query. This additional information enhances the accuracy of the VLLM's decision-making process. Consequently, evaluating visual queries with an API yields improved performance. Figure \ref{fig:visual_improve} supports this analysis by illustrating the performance gain achieved when using the API with a map image compared to relying solely on the map to solve the query with a VLLM.

Notably, in the routing category, Qwen-2.5-VL-72B exhibits a significant 24.46\% accuracy gain (as shown in Fig. \ref{fig:visual_improve}). This improvement is attributed to the difficulty in determining the optimal path by merely examining a map image. However, obtaining path information through route tools and calculating the optimal path based on time and distance is straightforward. This capability assists the VLLM in identifying the correct route. Listings \ref{lst:visual1} and \ref{lst:visual2} illustrate two scenarios where the response generated using MapAgent yields correct results, whereas the VLLM, without API assistance, fails to produce accurate outcomes.

\subsection{Why does API-using MapAgent perform better on contextual text data?}
\begin{figure}[t]
    \centering
    \includegraphics[width=0.9\linewidth]{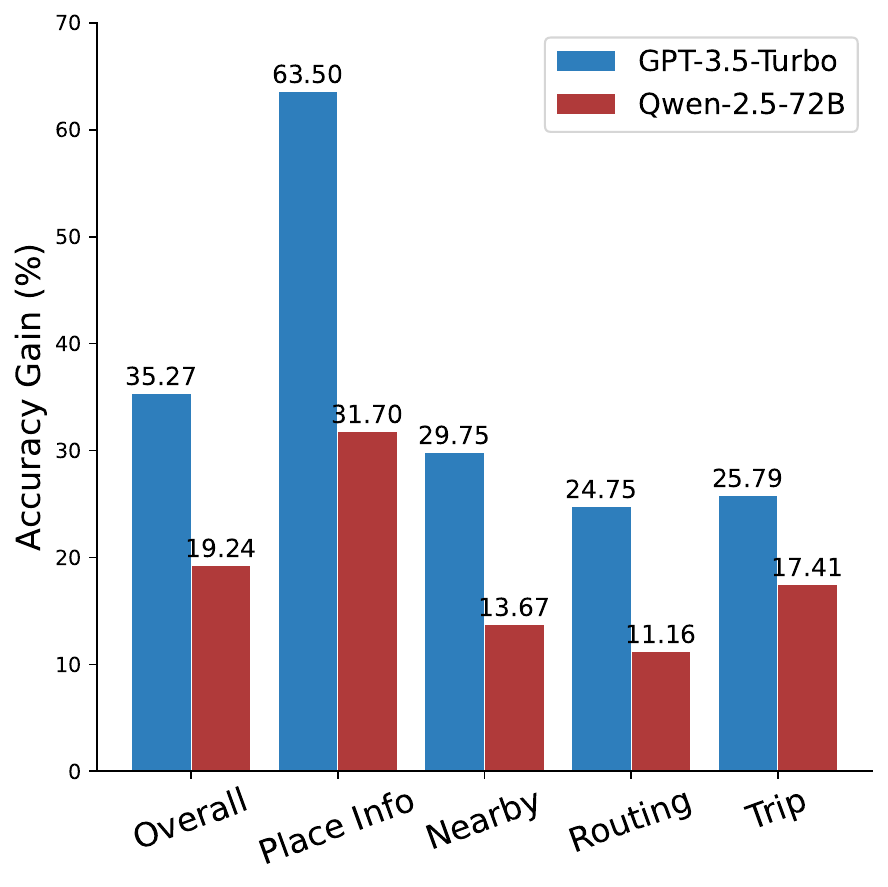}
    \caption{Performance gain from combining contextual text with API data over text alone.}
    \label{fig:textual_improve}
    \vspace{-10pt}
\end{figure}

Figure \ref{fig:textual_improve} illustrates the accuracy gain achieved by incorporating API calls with textual context. Notably, for the GPT-3.5-Turbo model, the overall gain is approximately 35.27\%, while for the Qwen-2.5-72b model, it is 19.24\%. The primary reason for this improvement is that the API can specifically retrieve information relevant to the question. In contrast, textual context may contain extraneous details, making it non-trivial to extract the precise information needed for an accurate answer. Furthermore, the context might lack essential information that can be obtained through the API. Therefore, to effectively address the query, the API plays a crucial role by providing supplementary relevant information. This additional input assists the LLM in predicting a more accurate response.

For qualitative examples illustrating agent behavior, please refer to Appendix~\ref{app:qualitative-examples}.

\section{Cost Analysis}
\label{cost-analysis}

\begin{figure}[h!]
  \centering
  \hspace{-12pt}
\includegraphics[width=1.04\linewidth]{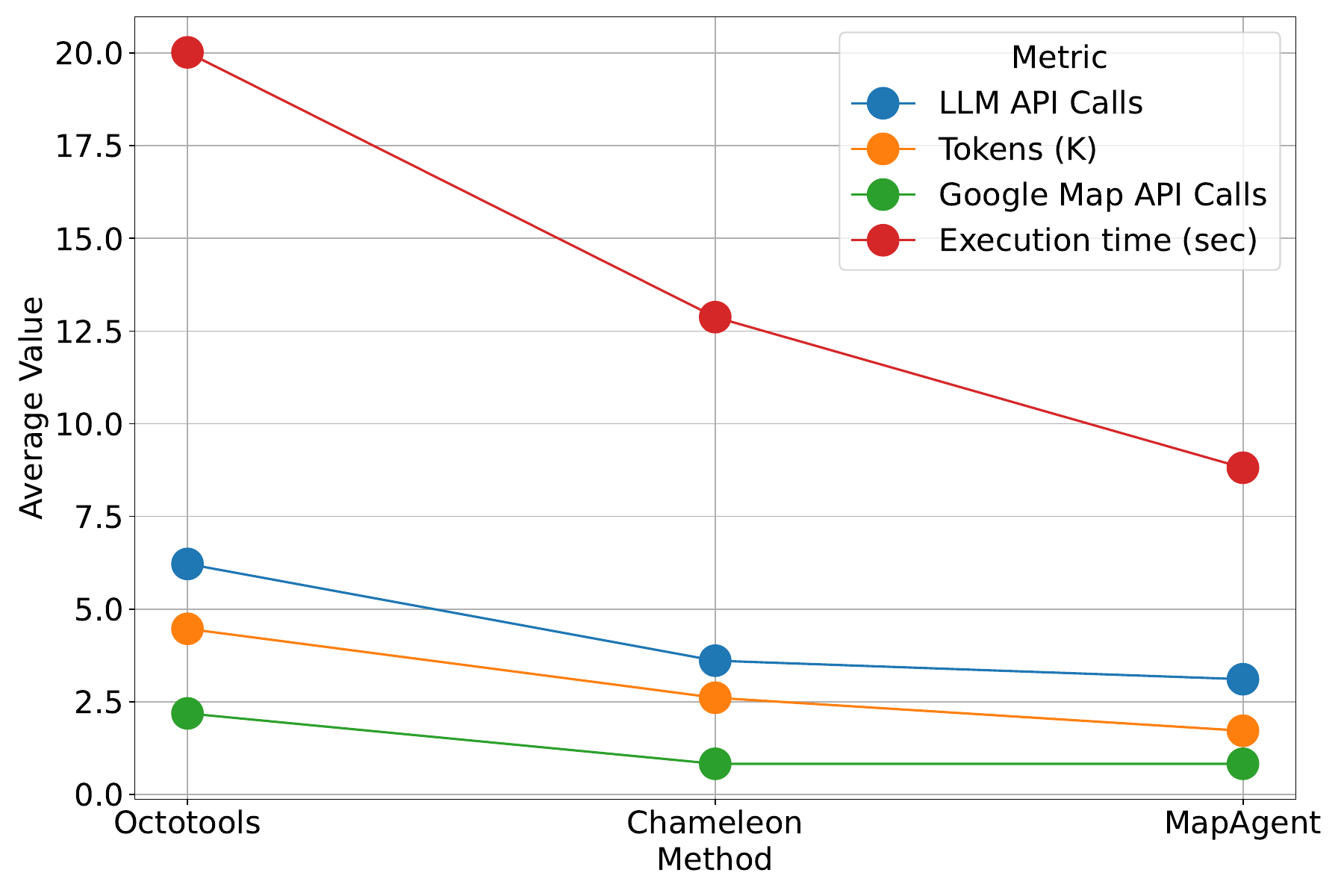}
\caption{Cost analysis of \mapagent{}compared with Chameleon and Octotools}
 \label{fig:cost-analysis-fig}
\end{figure}

For the cost analysis, we compare the costs associated with MapAgent to those of Chameleon and Octotools. We examine four key criteria: the LLM API call, the number of tokens used, the Google Maps API call, and the average execution time. As \Cref{fig:cost-analysis-fig} illustrates, MapAgent's cost consumption is lower when compared with Octotools and Chameleon. We observe that MapAgent achieves a shorter execution time with fewer LLM API calls and tokens used. Further statistical details regarding the cost analysis breakdown are provided in \Cref{cost_details}.

\begin{table*}[h]
\centering
\begin{tabular}[width=1\linewidth]
{|c|c|c|c|c|}
\hline
\multirow{2}{*}{\textbf{Dataset}} & \textbf{Planner Agent} & \multicolumn{2}{|c|}{\textbf{Map-Tool Agent}} & \multirow{2}{*}{\textbf{Overall Error(\%)}} \\
\cline{3-4}
& \textbf{Module Selection} & \textbf{Tools Calling} & \textbf{Parameter Passing} &  \\
\hline
\apitool & 1.33 & 0.76 & 1.46 & 3.55\\
\hline
\textualtool & 0.67 & 1.02 & 0.93 & 2.62\\
\hline
MapQA & 0.89 & 1.56 & 1.78 & 4.23\\
\hline
\visualtool & 2.5 & 0.66 & 1.2 & 4.36 \\
\hline
\end{tabular}
\caption{\mapagent{} Error(\%) Analysis}
\label{tab:error-analysis}
\end{table*}

\section{Error Analysis and Challenges }
\label{error-analysis}

\begin{figure}[h]
\vspace{-4pt}
    \centering
     \includegraphics[width=1\linewidth]{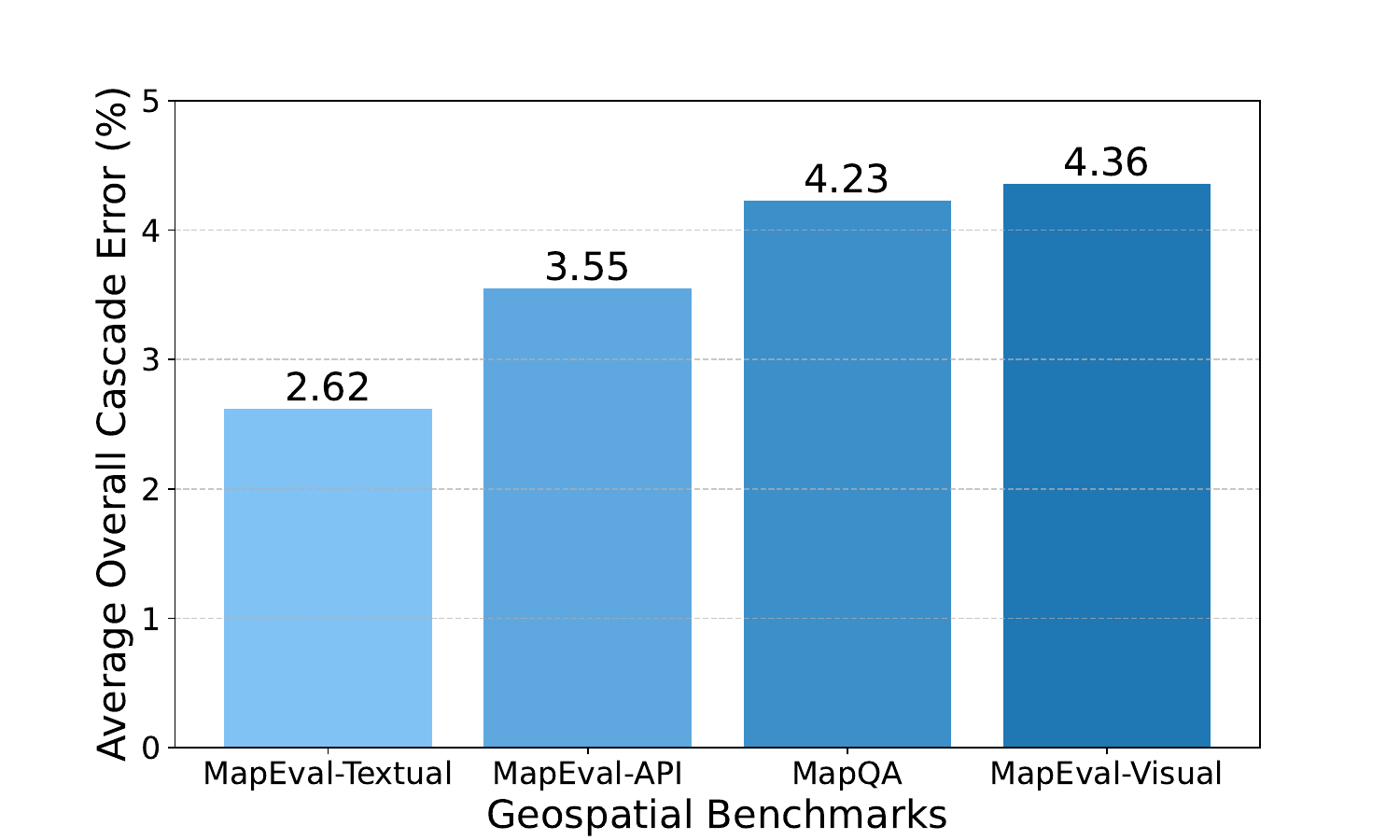}
    \caption{Average Overall Error(\%) in MapAgent across 4 geo-spatial benchmarks.}
    \label{fig:error-analysis-fig}
\end{figure}

To pinpoint the remaining challenges of MapAgent, we conducted a granular error analysis, examining specific error types at various stages of the framework. As illustrated in Figure \ref{fig:error-analysis-fig}, MapAgent exhibits an average error percentage across four geospatial benchmarks. While the average error rate is not significant, it stands at approximately 4.36\% for MapEval-Visual datasets, 4.24\% for MapQA, 3.55\% for MapEval-API, and 2.62\% for MapEval-Textual.
Furthermore, Table \ref{tab:error-analysis} details the average percentage of errors attributed to each component within MapAgent. Our analysis reveals several key error sources:

\noindent \textbf{Planner Agent:} While generally accurate, the Planner Agent occasionally fails to select the appropriate modules from its inventory. We observed varying error rates depending on the task: approximately 0.67\% for MapEval-Textual, 0.89\% for MapQA, 1.33\% for MapEval-API, and 2.5\% for MapEval-Visual. These errors often arise from wrong component choices or operation order.

\noindent \textbf{Map-Tool Agent:} This agent can encounter two potential types of errors: fetching incorrect tools or passing improper arguments to those tools. While the error rate is not significant, it is detailed in \Cref{tab:error-analysis}.

\section{Conclusion}
\label{sec:conclusion}

In this paper, we present \mapagentnospace, a hierarchical multi-agent framework designed for effective geospatial reasoning. By decoupling planning from execution and introducing specialized modules with customized toolset--such as a dedicated map-tool agent--\mapagent addresses the limitations of flat agent-based architectures that struggle with tool overload and fine-grained API coordination. Through comprehensive evaluations on four challenging geospatial benchmarks—MapEval-Textual, MapEval-API, MapEval-Visual, and MapQA-\mapagent consistently outperforms existing tool-augmented and agentic baselines, including Chameleon and OctoTools. These results demonstrate \mapagent’s effectiveness and generalizability in real-world, map-integrated reasoning tasks. In future work, we plan to extend the framework to support multi-agent collaboration \cite{hosain2025xolver} and to address broader multimodal spatio-temporal reasoning tasks.

\section*{Limitations}
While our proposed hierarchical multi-agent framework, \mapagentnospace, demonstrates substantial gains over state-of-the-art tool-augmented and agentic baselines for a variety of geospatial queries, it currently relies on a limited set of popular Google Maps APIs. Extending the framework to support a broader range of APIs and mapping services would further validate its generalizability and robustness. Additionally, although \mapagent features a plug-and-play architecture with the potential to be applied to other domains, such as web automation and software engineering, its effectiveness in these areas remains to be evaluated.

\bibliography{custom}

\clearpage
\newpage
\appendix
\section{How \mapagent Outperforms Chameleon and OctoTools}
Appendices \ref{app:octotools}, \ref{app:chameleon}, and \ref{app:mapagent} illustrate an example where OctoTools and Chameleon failed to solve a complex geospatial query, but MapAgent could. The main reason behind this failure is that both OctoTools and Chameleon have all their module or tool cards in one layer. Consequently, they have trouble guessing the exact tool/module for fetching information. The example illustrates that solving this query requires the distance and travel time from one location to another for each location in the query, which a trip tool can deliver correctly. However, both Chameleon and OctoTools struggle here and select a route tool to fetch the information, subsequently lacking information about all the routes between all the places. Additionally, OctoTools encounters another limitation. As shown in Table \ref{tab:reasoning_example_1}, where the intermediate reasoning steps of OctoTools are displayed, we found that this framework hallucinates and triggers a tool with parameters that have no relation to this question, which yields wrong information. These limitations trigger a failure to provide all the required information to the subsequent steps. However, MapAgent can effectively solve this by introducing an inner layer in the Google Map module. This module, communicating with a map-tool agent, appropriately triggers the trips tool and retrieves the relevant information that is sufficient for the subsequent steps to determine the solutions.

\begin{figure}[h!]
    \centering
    \includegraphics[width=1\linewidth]{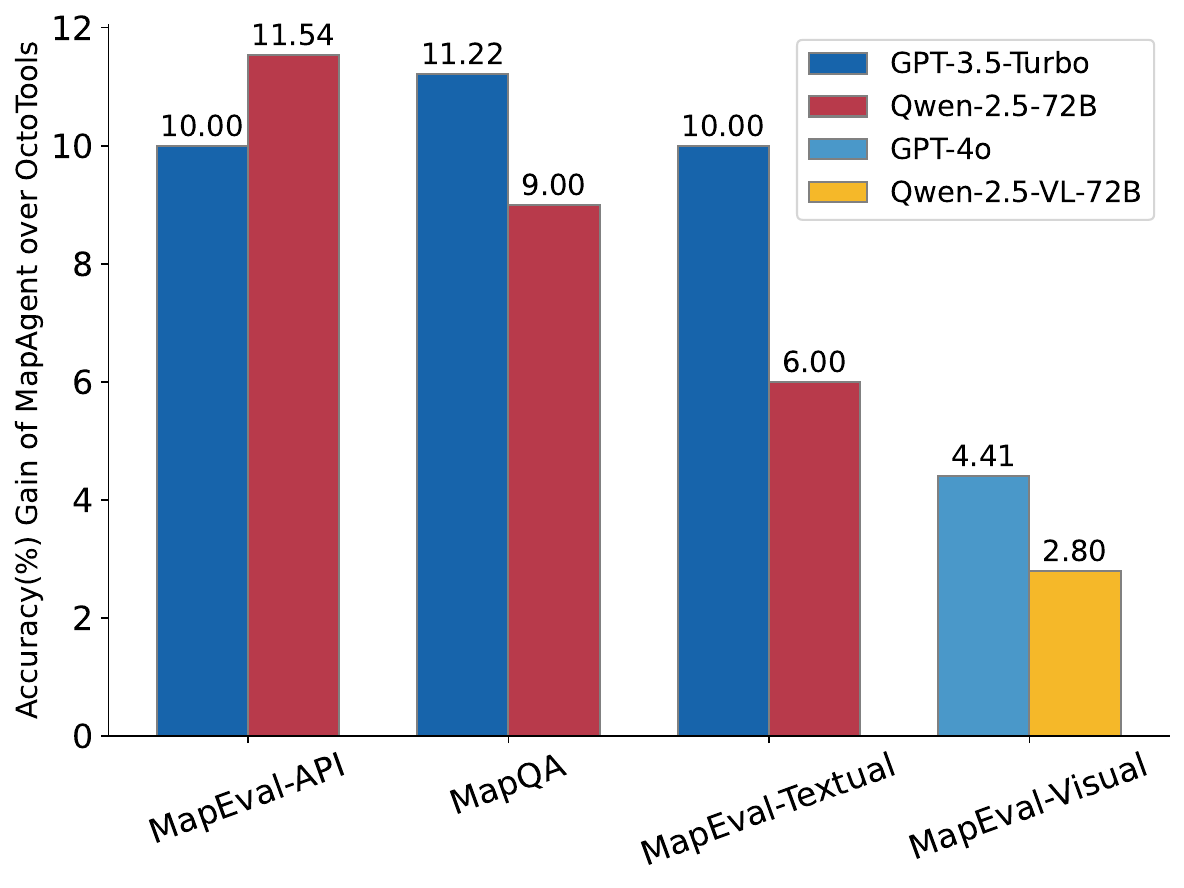}
    \caption{Accuracy improvements of MapAgent over OctoTools on various benchmarks.}
    \label{fig:mapagent_vs_octotools}
\end{figure}

\subsection{Octotools Output \colorbox{green}{Green:}Correct Answer \colorbox{red}{Red} Wrong Answer}
\label{app:octotools}
\begin{itemize}
    \item \textbf{PID:} 19
    \item \textbf{Query:} I am at Rydges Wellington. Starting from my current location, I will first go to  Kelburn Park, followed by a trip to Victoria University of Wellington, and finally return to my current location. All the travel will be done using my own car. How much time will the entire journey take? 
    \item \textbf{Options:} A.11 mins  B.10 mins  C.12 mins  D.9 mins
    \item \textbf{Correct Answer:} B
    \item \textbf{Reasoning Steps:} Table \ref{tab:reasoning_example_1} shows the reasoning steps of OctoTools and the tools it executes in each step. Within the table, \colorbox{red}{red} highlights indicate hallucinations, and \colorbox{orange}{orange} highlights denote incorrect information that OctoTools attempts to fetch after execution. We observe that the initial call to the Route Tool, using the same origin and destination, leads to inaccurate information retrieval. Furthermore, the subsequent nine calls, shown in table rows 2 through 5, demonstrate hallucination and the use of parameters unrelated to the question, revealing a lack of correct tool calling. Although a trip tool capable of providing all relevant information in a single call exists, it is not executed correctly.
    \item \textbf{Octotools Final Output:} To determine the total travel time for the round trip starting and ending at Rydges Wellington, with stops at Kelburn Park and Victoria University of Wellington, the following steps were taken:
    
    Step 1: Analyze the Query
    The query specifies a round trip:
    1. Start at **Rydges Wellington**.
    2. Visit **Kelburn Park**.
    3. Proceed to **Victoria University of Wellington**.
    4. Return to **Rydges Wellington**.
    The mode of transport is driving.
    
    Step 2: Use the Route Tool
    The **Route Tool** was used to calculate the total driving time for the specified route. The tool provided the following result:
    - **Total Route Distance**: 1.9 km.
    - **Total Driving Time**: 9 minutes.
    
     Step 3: Verify the Route Details
    The route details confirm the driving time includes all segments of the trip:
    - Rydges Wellington → Kelburn Park.
    - Kelburn Park → Victoria University of Wellington.
    - Victoria University of Wellington → Rydges Wellington.
    
    Conclusion:
    The total driving time for the entire journey is **9 minutes**.

    **Final Answer: \colorbox{red}{D. 9 mins**}

\end{itemize}

\begin{table*}[h]
    \centering
    \caption{Detailed Reasoning Steps of Octotools}
    \label{tab:reasoning_example_1}
    \begin{tabular}{p{1cm} p{1cm} p{4cm} p{5cm} p{3cm}}
        \toprule
        
        \textbf{Step} & \textbf{Tool} & \textbf{Sub-goal} & \textbf{Command} & \textbf{Result Summary} \\
        \midrule
        Action Step 1 & Route Tool & Calculate the total driving time for the round trip starting at Rydges Wellington, visiting Kelburn Park and Victoria University of Wellington, and returning to Rydges Wellington. & \cellcolor{orange}\texttt{\seqsplit{execution = tool.execute(origin="Rydges Wellington", destination="Rydges Wellington", waypoints=["Kelburn Park", "Victoria University of Wellington"], mode="driving", alternatives=False)}} & \cellcolor{orange}Total 1 route. VIA Dixon St (1.9 km, 9 mins). Details include step-by-step directions. \\
        \rowcolor{red}
        Action Step 2 & Route Tool & Calculate the travel time from the Trevi Fountain to the Vatican Museum using public transport. & \texttt{\seqsplit{execution = tool.execute(origin="Trevi Fountain", destination="Vatican Museum", mode="transit")}} & Total 6 routes from Trevi Fountain to Vatican Museum via different public transport options with times ranging from 30 to 39 mins. \\
        \rowcolor{red}
        Action Step 3 & Route Tool & Calculate the travel time for the transit leg from Kockelscheuer Ice Rink to Kallektuffquell using the Route\_Tool. & \texttt{\seqsplit{execution = tool.execute(origin="Kockelscheuer Ice Rink", destination="Kallektuffquell", mode="transit")}} & Total 6 routes from Kockelscheuer Ice Rink to Kallektuffquell via public transport with times ranging from 1 hour 50 mins to 2 hours 10 mins. \\
        \rowcolor{red}
        Action Step 4 & Route Tool & Calculate the total driving time for the route from Planetshakers Church Melbourne City Campus to Melbourne Lifestyle Apartments - Docklands Accommodation with a stop at Carrot Tattoo Studio & \texttt{\seqsplit{execution = tool.execute(origin="Planetshakers Church Melbourne City Campus", destination="Melbourne Lifestyle Apartments - Docklands Accommodation", waypoints=["Carrot Tattoo Studio"], mode="driving", alternatives=False)}} & Total 1 route from Planetshakers Church to Melbourne Lifestyle Apartments (via Carrot Tattoo Studio) (2.0 km, 9 mins) with step-by-step directions. \\
        \rowcolor{red}
        Action Steps 5-10 & Route Tool & Calculate the total driving time for the route from Hotel Saltel to Nil Ratan Sircar Medical College and Hospital with a stop at Science City - Kolkata. & \texttt{\seqsplit{execution = tool.execute(origin="Hotel Saltel", destination="Nil Ratan Sircar Medical College and Hospital", waypoints=["Science City - Kolkata"], mode="driving", alternatives=False)}} & Each step shows 1 route from Hotel Saltel to Nil Ratan Sircar Medical College (via Science City - Kolkata) (2.8 km, 7 mins) with step-by-step directions. \\
        \bottomrule
    \end{tabular}
\end{table*}
\subsection{Chameleon Output \colorbox{green}{Green:}Correct Answer \colorbox{red}{Red} Wrong Answer}
\label{app:chameleon}
\begin{itemize}
    \item \textbf{PID:} 19
    \item \textbf{Query:} I am at Rydges Wellington. Starting from my current location, I will first go to  Kelburn Park, followed by a trip to Victoria University of Wellington, and finally return to my current location. All the travel will be done using my own car. How much time will the entire journey take? 
    \item \textbf{Options:} A.11 mins  B.10 mins  C.12 mins  D.9 mins
    \item \textbf{Correct Answer:} B
    \item \textbf{Selected Module by the Planner:} Route, solution\_generator, answer\_generator
    \item \textbf{Information gathered by Route Module:} There are total 1 routes from Rydges Wellington, Wellington 6011, New Zealand to Victoria University of Wellington, Wellington 6012, New Zealand. The route information is provided below:\\
     Route 1:(VIA) Parkvale Rd (7.6 km, 20 mins). Details steps are provided below: \\
     Head southwest on Featherston St toward Whitmore St (35 m, 1 min) \\
     Turn right at the 1st cross street onto Whitmore St (0.2 km, 1 min) \\
     Continue onto Bowen St (0.8 km, 2 mins) \\
     Turn left onto Tinakori Rd (signs for Karori/Botanic Garden) (0.2 km, 1 min) \\
     Continue onto Glenmore St (1.6 km, 3 mins) \\
     At the roundabout, take the 1st exit and stay on Glenmore St (0.2 km, 1 min) \\
     Continue onto Karori Tunnel (0.1 km, 1 min) \\
     Continue onto Chaytor St (0.9 km, 2 mins) \\
     Continue onto Karori Rd (1.2 km, 2 mins) \\
     Turn right onto Parkvale Rd (1.1 km, 2 mins) \\
     Turn right to stay on Parkvale Rd (0.3 km, 1 min) \\ 
     Turn right to stay on Parkvale Rd (1.1 km, 5 mins)
    \item \textbf{Solution Generated by Solution Generator:} It seems there's a mismatch in the information provided. The detailed route you've given is for travel between Rydges Wellington and Victoria University of Wellington, taking approximately 20 minutes. You haven't provided a detailed route or travel time for the leg between Rydges Wellington and Kelburn Park, or for the return trip from Victoria University of Wellington back to Rydges Wellington via Kelburn Park.
     Based solely on the 20-minute travel time you just provided for Rydges to Victoria University (Route 1), we cannot accurately answer the question about the entire journey involving Kelburn Park without knowing the travel times for the other segments. 
    However, if we were to ignore the previous travel times you mentioned and assume the question implies a direct trip from Rydges to Victoria University and back, using the 20-minute route in each direction, the total would be 20 minutes + 20 minutes = 40 minutes. This doesn't align with the provided options. So, the answer should be E.
    \item \textbf{Answer Generator Module Output:} \colorbox{red}{Not able to answer the question}

\end{itemize}

\subsection{\mapagent Output \colorbox{green}{Green:}Correct Answer \colorbox{red}{Red} Wrong Answer}
\label{app:mapagent}
\begin{itemize}
    \item \textbf{PID:} 19
    \item \textbf{Query:} I am at Rydges Wellington. Starting from my current location, I will first go to  Kelburn Park, followed by a trip to Victoria University of Wellington, and finally return to my current location. All the travel will be done using my own car. How much time will the entire journey take? 
    \item \textbf{Options:} A.11 mins  B.10 mins  C.12 mins  D.9 mins
    \item \textbf{Correct Answer:} B
    \item \textbf{Selected Module by the Planner Agent:} map\_service, solution\_generator, answer\_generator
    \item \textbf{Information gathered by Google\_Maps Module:} \\
    All Location Info: \\
    Rydges Wellington \\
    Name: Rydges Wellington \\
    Address: 75 Featherston Street, Pipitea, Wellington 6011, New Zealand \\
    Rating: 4.2 \\
    Types: lodging, point\_of\_interest, establishment \\
    Is Open Now: True \\
    Weekday Opening Hours: \\
    - Monday: Open 24 hours \\
    - Tuesday: Open 24 hours \\
    - Wednesday: Open 24 hours \\
    - Thursday: Open 24 hours \\
    - Friday: Open 24 hours \\
    - Saturday: Open 24 hours \\
    - Sunday: Open 24 hours \\
    Kelburn Park \\
    Name: Kelburn Park \\
    Address: Salamanca Road, Kelburn, Wellington 6012, New Zealand \\
    Rating: 4.4 \\
    Types: park, tourist\_attraction, point\_of\_interest, establishment \\
    Is Open Now: True \\
    Weekday Opening Hours: \\
    - Monday: Open 24 hours \\
    - Tuesday: Open 24 hours \\
    - Wednesday: Open 24 hours \\
    - Thursday: Open 24 hours \\
    - Friday: Open 24 hours \\
    - Saturday: Open 24 hours \\
    - Sunday: Open 24 hours \\
    Victoria University of Wellington \\
    Name: Victoria University of Wellington \\
    Address: Kelburn Parade, Kelburn, Wellington 6012, New Zealand \\
    Rating: 4.3 \\
    Types: university, point\_of\_interest, establishment \\
    Is Open Now: N/A \\
    Weekday Opening Hours: \\
    - Unknown \\
    The travel time(distance) from Rydges Wellington to Kelburn Park is 5 mins (1.6 km) \\
    The travel time(distance) from Rydges Wellington to Victoria University of Wellington is 1 min (1 m) \\
    The travel time(distance) from Kelburn Park to Rydges Wellington is 4 mins (1.4 km) \\
    The travel time(distance) from Kelburn Park to Victoria University of Wellington is 4 mins (1.4 km) \\
    The travel time(distance) from Victoria University of Wellington to Rydges Wellington is 1 min (1 m) \\
    The travel time(distance) from Victoria University of Wellington to Kelburn Park is 5 mins (1.6 km)
    \item \textbf{Solution Generated by Solution Generator Module:} To calculate the total time for the entire journey, we need to consider the travel time from Rydges Wellington to Kelburn Park, then from Kelburn Park to Victoria University of Wellington, and finally the return trip from Victoria University of Wellington to Rydges Wellington.

    The travel time from Rydges Wellington to Kelburn Park is 5 minutes.
    The travel time from Kelburn Park to Victoria University of Wellington is 4 minutes.
    The return trip from Victoria University of Wellington to Rydges Wellington is 1 minutes.
    
    Adding these times together, the total time for the entire journey is 5 + 4 + 1 = 10 minutes.\\
    So, the answer is 10 minutes. The most possible option is "the answer is B
    
    \item \textbf{Answer Generated by Answer Generator Module:} \colorbox{green}{B. 10 mins}

\end{itemize}

\begin{table*}[t]
\centering
\begin{tabular}{>{\raggedright\arraybackslash}p{3.2cm}|p{11.8cm}}
\hline
\textbf{Module} & \textbf{Description} \\
\hline
\multirow{3}{*}{\textbf{Map Service}} & Fetches geospatial data using a Coordinator Agent called Map-Tool Agent. Retrieves place details, routes, nearby locations, and other map-related information. \\
\hline
\textbf{Visual Place Recognizer } & Uses VLLM to analyze geospatial imagery. Identifies the central geographic
                      place and estimates the geographic scope (approximate radius). \\
\hline
\multirow{2}{*}{\textbf{Sequencer}} & Organizes and structures the responses from preceding modules into a logical sequence. \\
\hline
\multirow{2}{*}{\textbf{Solution Generator}} & Uses a LLM or VLLM to synthesize the final answer based on structured Input. \\
\hline
\multirow{2}{*}{\textbf{Answer Generator}} & Refines and verifies the solution produced by the Solution Generator. Extracts and checks the predicted answer for consistency. \\
\hline
\end{tabular}
\caption{Modules in the Module Inventory essential for solving geospatial queries}
\label{tab:modules}
\end{table*}

\begin{table*}[h]
\centering
\begin{tabular}{p{3cm}|p{11.5cm}}
\hline
\textbf{Tool} & \textbf{Description} \\
\hline
\textbf{Trip Tool} & Retrieves information about locations and travel routes between them. Combines the Place Details and Directions APIs to fetch place metadata and step-by-step route instructions. \\
\hline
\textbf{Route Tool} & Fetches route information between two points using the Directions API, including distance, estimated time, and navigation steps. \\
\hline
\textbf{Nearby Tool} & Identifies points of interest within a specified area using the Nearby Search API. Returns place names, ratings, and other metadata. \\
\hline
\textbf{PlaceInfo Tool} & Retrieves detailed information about a given place using the Place Details API, including address, hours, contact info, and reviews. \\
\hline
\end{tabular}
\caption{Map tools used by the \mapagent{} framework, each built over a specific Google Maps API.}
\label{table:tools}
\end{table*}

\section{Configurations in MapAgent}

\subsection{Map-Tool Agent}

\begin{tcolorbox}[title=Prompt for Map-Tool Agent, colback=blue!5!white, colframe=blue!75!black]
You are an agent designed to understand user questions and retrieve relevant information using specific tools. When a user asks a question, your task is to identify the appropriate tool from the available list (Trip Tool, Route Tool, Nearby, and PlaceInfo) that can best answer the query. You will then use that tool to fetch the information and provide the retrieved data to the user. You are not expected to answer the question directly; your role is solely to retrieve the necessary information using the designated tools. For instance, if you think you need to use more than one tool, you can use them in parallel.
\end{tcolorbox}

\subsection{Solution Generator (MapEval-API)}
\begin{tcolorbox}[title=Prompt for Solution Generator, colback=purple!5!white, colframe=purple!75!black]
Given the question (and the context), select the answer from the options ["A", "B", "C", "D"]. You should give concise and step-by-step solutions. Finally, conclude the answer in the format of "the answer is [ANSWER]", where [ANSWER] is one from the options ["A", "B", "C", "D"]. For example, "the answer is A", "the answer is B", "the answer is C", or "the answer is D". If the answer is not in the options, declare it as Unanswerable.
\\
\textbf{\# Example 1}

\textbf{Question:} What is the address of Multiplan Center?

\textbf{Options:} (A) 69, 71 New Elephant Rd, Dhaka 1205, Bangladesh  (B) 38/1/C BC DAS Street Lalbagh  (C) Polashi,BUET  (D) Central Road, USA

\textbf{Metadata:} \{"skill":"Fetch context from corresponding google map api and based on the context answer the question"\}

\textbf{Google Maps response:}

Name: Multiplan Center\\
Address: 69, 71 New Elephant Rd, Dhaka 1205, Bangladesh\\
Rating: 4.4\\
Types: point\_of\_interest, establishment\\
Is Open Now: False\\
Weekday Opening Hours:\\
- Monday: 10:00AM–8:00PM\\
- Tuesday: Closed\\
- Wednesday: 10:00AM–8:00PM\\
- Thursday: 10:00AM–8:00PM\\
- Friday: 10:00AM–8:00PM\\
- Saturday: 10:00AM–8:00PM\\
- Sunday: 10:00AM–8:00PM\\
\textbf{Solution:} If you look at the context and search, then after reaching The address of Multiplan Center is 69, 71 New Elephant Rd, Dhaka 1205, Bangladesh. Therefore, the answer is B.\\

.... <4 more examples> ....\\

Now Answer the question following.

\textbf{\# Question:} \{\texttt{question}\}
\end{tcolorbox}

\subsection{Visual Place Recognizer}

\begin{tcolorbox}[title=Prompt for Visual Place Recognizer, colback=green!5!white, colframe=green!75!black]
I will provide you with an image. You must determine the precise center location within the image and a prediction of the boundary of the image by predicting a radius from the center. Your return format should be the center location name followed by a space, then the radius. Based on that, provide only one complete address in a single, consistent format followed by a space and a radius. Do not include any extraneous text before or after the address.
\end{tcolorbox}
\section{Detailed Tool Parameters}
\label{appendix:detail_table_parameter}

The following section outlines the tools utilized by the Map Service module, as retrieved by the map-tool agent. Each tool is defined along with its input parameters.

\begin{tcolorbox}[title=Trip Tool, colback=blue!5!white, colframe=blue!75!black]
\texttt{\textbf{current\_location} (str)}: The starting location of the trip. \\
\texttt{\textbf{visiting\_places} (list)}: A list of locations to visit. \\
\texttt{\textbf{travel\_mode} (str)}: The mode of travel, defaults to driving. \\
\end{tcolorbox}

\begin{tcolorbox}[title=Route Tool, colback=blue!5!white, colframe=blue!75!black]
\texttt{\textbf{origin} (str)}: The starting location for the directions. \\
\texttt{\textbf{destination} (str)}: The destination for the directions. \\
\texttt{\textbf{travel\_mode} (str)}: The mode of travel, defaults to driving. \\
\texttt{\textbf{mode} (str)}: The mode of transportation to use for the directions, such as "driving" \\
\texttt{\textbf{alternatives} (bool)}: Whether to return multiple possible routes. 
\end{tcolorbox}

\begin{tcolorbox}[title=Nearby Tool, colback=blue!5!white, colframe=blue!75!black]
\texttt{\textbf{query} (str)}: The search term to look for geospatial places. \\
\texttt{\textbf{location} (str)}: The name of the current location (e.g., Ibn Sina Hospital, Dhaka). \\
\texttt{\textbf{type} (str)}: The type of place to search for, such as "restaurant", "cafe", or "hospital". \\
\texttt{\textbf{radius} (float)}: The radius of the circular area for filtering, defaults to 20 Kilometer  \\
\end{tcolorbox}

\begin{tcolorbox}[title=PlaceInfo Tool, colback=blue!5!white, colframe=blue!75!black]
\texttt{\textbf{location\_address} (str)}: The address or name of the location to search for. \\
\end{tcolorbox}

\section{Details Cost Analysis}
\label{cost_details}

The cost analysis reveals that MapAgent is consistently more efficient and cost-effective than Chameleon and Octotools across all tested LLM models and datasets. This is due to MapAgent's superior performance in four key areas: execution time, LLM API calls, Google Map API calls, and token consumption, as shown in Figures \cref{fig:execution-time}, \Cref{fig:llm-api-calls}, \Cref{fig:token}, and \Cref{fig:google Maps api}, respectively.

\begin{table*}[h]
\centering
\includegraphics[width=1\linewidth]{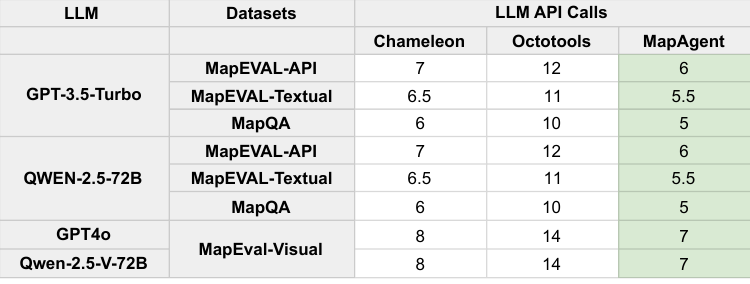}
\caption{Average LLM API Calls}
\label{fig:llm-api-calls}
\end{table*}

\begin{table*}[h]
\centering
\includegraphics[width=1\linewidth]{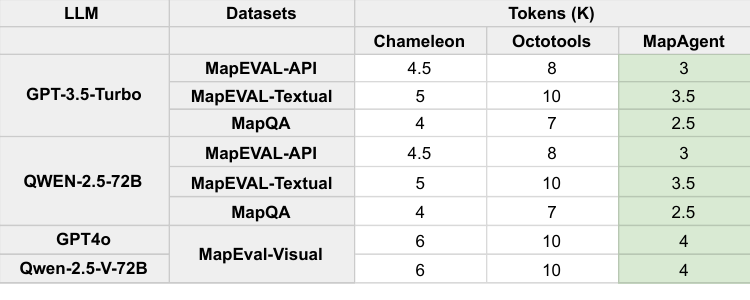}
 \caption{Average Tokens Required}
\label{fig:token}
\end{table*}

\begin{table*}[h]
\centering
\includegraphics[width=1\linewidth]{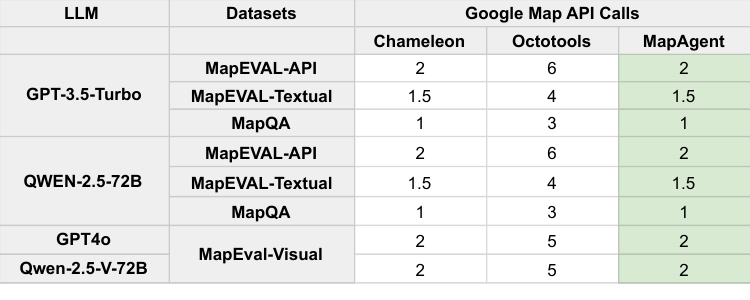}
\caption{Average Google Maps API Calls}
\label{fig:google Maps api}
\end{table*}

\begin{table*}[h]
\centering
\includegraphics[width=1\linewidth]{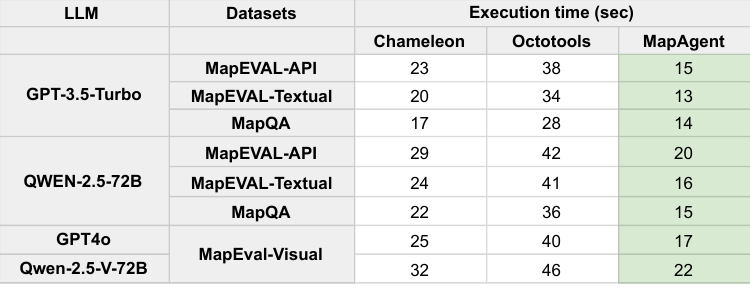}
\caption{Average Execution Time}
\label{fig:execution-time}
\end{table*}

\section{Qualitative Examples}\label{app:qualitative-examples}
\subsection{How \mapagent Answer a Query}
\begin{figure*}[h]
\centering
\includegraphics[width=1.0\linewidth]{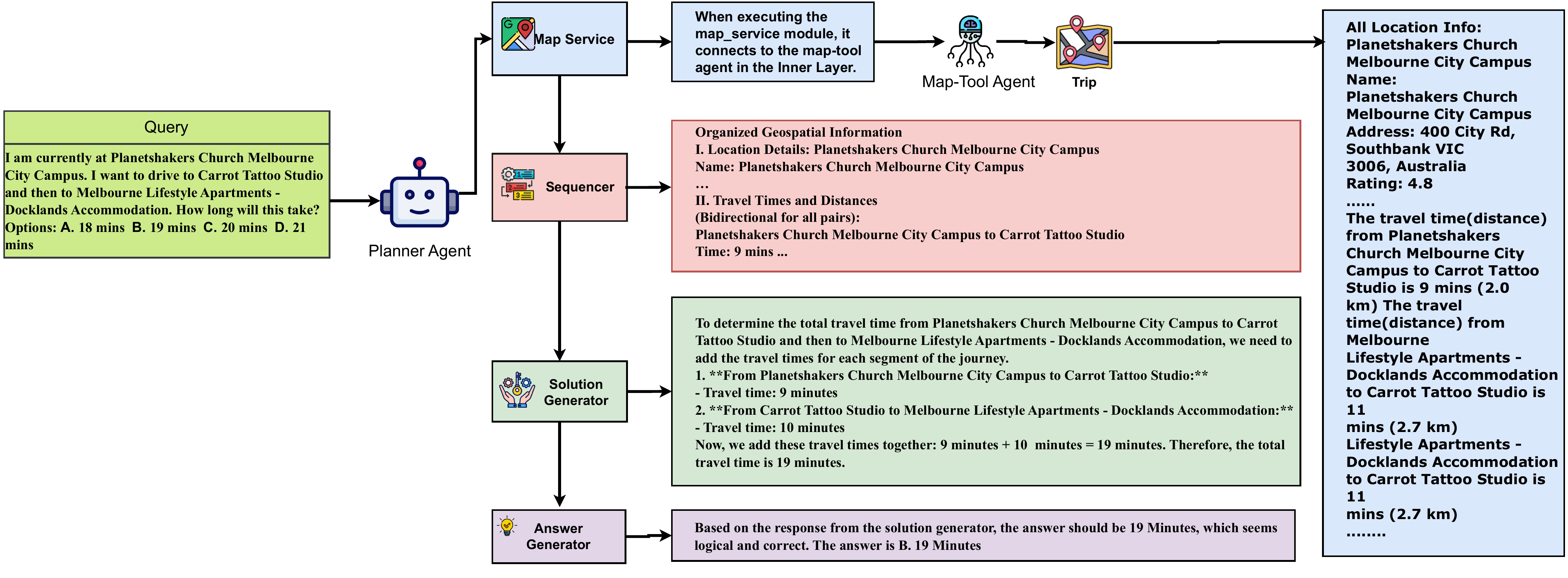}
\caption{Detailed Example: How MapAgent Solves a Query, Including the Output of Each Module}
\vspace{-10pt}
\label{fig:example_1}
\end{figure*}

Listing \ref{lst:straight-ex} and \Cref{fig:example_1} illustrates how a \mapagent addresses a typical geospatial query. In this example, the query is taken from the trip category: "\textit{I am currently at Planetshakers Church Melbourne City Campus. I want to drive to Carrot Tattoo Studio and then to Melbourne Lifestyle Apartments - Docklands Accommodation. How long will this take?}"

Initially, the query is passed to the Planner Agent, which analyzes it and selects three modules to solve it: google\_maps, solution\_generator, and answer\_generator. Subsequently, the google\_maps module is executed and attempts to fetch information from Google Maps by engaging its map-tool agent and relevant tools. The execution is then handed over to the map-tool agent. Based on the query, the map-tool agent selects the trip tool to retrieve the necessary information for this specific question. The trip tool utilizes the Directions and Place Details APIs to fetch information for solving this particular query. The map-tool agent then returns this information to the google\_maps module. The Google Maps module then passes this information to the solution\_generator. Based on the information fetched by google\_maps and the query itself, the solution\_generator attempts to formulate the answer with a proper explanation and hands this over to the answer\_generator. Finally, after verifying the answer, the answer\_generator extracts the final answer from the explanation provided by the solution\_generator and presents the final output. By decomposing the task into multiple steps, this approach addresses the long-context processing limitations of LLMs \cite{parvez-2025-chain}.

\subsection{How \mapagent Answer a Query with Visual Context}
\begin{figure*}[h]
\centering
\includegraphics[width=1.0\linewidth]{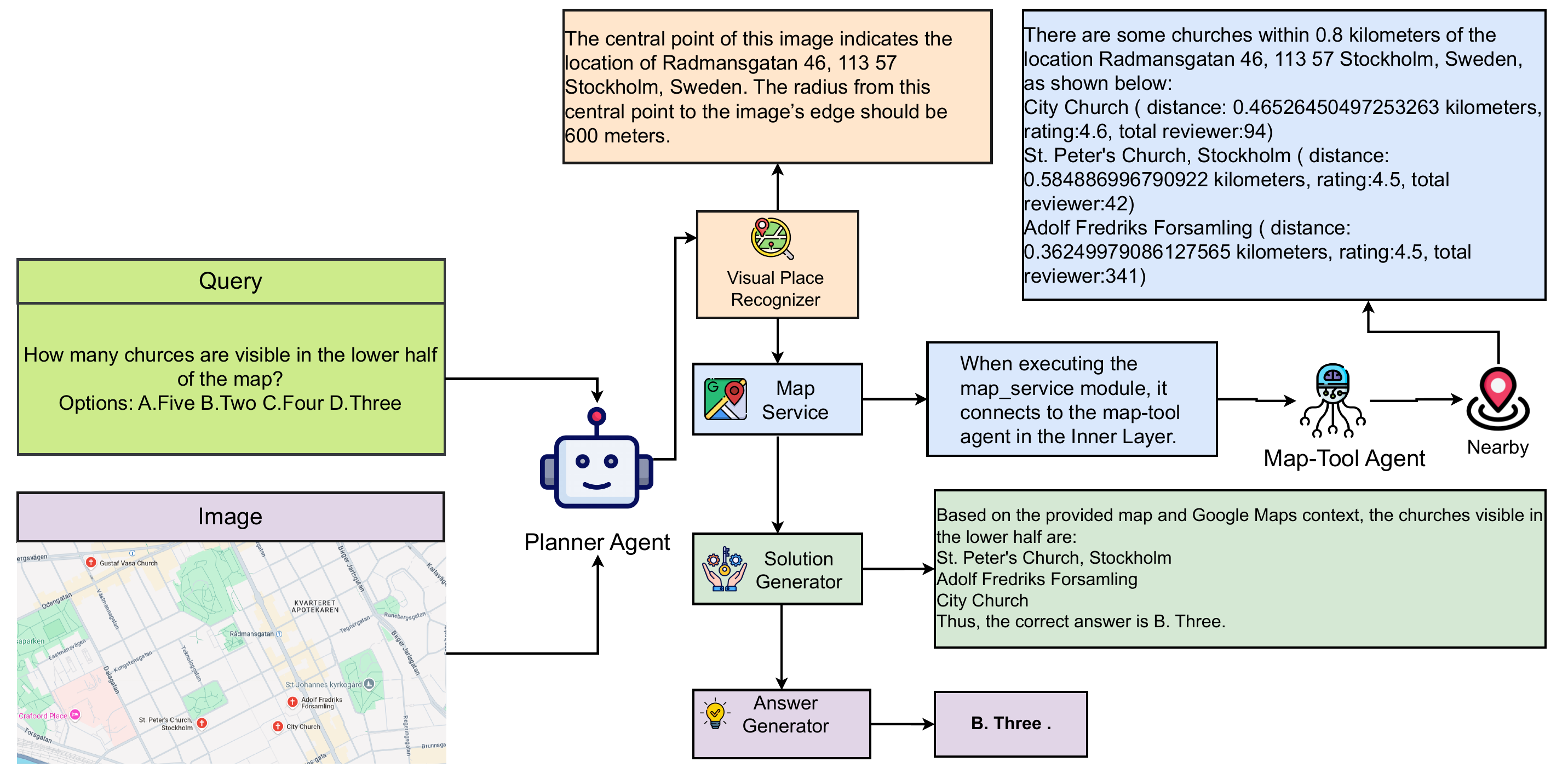}
\caption{Detailed Example: How MapAgent Solves a Query, Including the Output of Each Module}
\vspace{-10pt}
\label{fig:example_2}
\end{figure*}

Listings \ref{lst:visual-exm-1}, \ref{lst:visual-exm-2}, \ref{lst:visual-exm-2-cntd}, \ref{lst:visual-exm-3} and \Cref{fig:example_2} illustrate examples of how MapAgent handles geospatial queries that include a visual context, such as a snapshot of a location. To address this type of problem, the question and image are processed by the Planner Agent. The Planner Agent typically selects the Visual Place Recognizer Module, the \texttt{google\_maps} module, the \texttt{solution\_generator} module, and the \texttt{answer\_generator} module. The Image Search module identifies the central location within the image and estimates its boundaries, often providing a radius from the central location. This location information, along with the original query, is then passed to the \texttt{google\_maps} module. Google Maps, through its map-tool agent and tools that invoke the Google Maps API, fetches further relevant information from its inner layer. Finally, similar to the previous example, the solution is generated and refined. 

\begin{figure*}[h]
\begin{lstlisting}[caption={Example how \mapagent answer a query \colorbox{green}{Green}: Correct Answer. \colorbox{red}{Red}: Wrong Answer.},captionpos=t, basicstyle=\scriptsize\ttfamily, escapeinside={@}{@}, literate={hrule}{{\hrulefill}}5, label=lst:straight-ex, breaklines=true, breakindent=0pt, xleftmargin=0pt, frame=lines]
@\colorbox{lightblue}{Question:}@ 
I am currently at Planetshakers Church Melbourne City Campus. I want to drive to Carrot Tattoo Studio and then to Melbourne Lifestyle Apartments - Docklands Accommodation. How long will this take?
@\colorbox{yellow}{Options:}@ A. 18 mins @\colorbox{green}{B. 19 mins}@ C. 20 mins D. 21 mins
hrule
@\colorbox{lightgray}{Modules selected by \textbf{Planner agent}:}@ 
1.Map Service
2.Solution Generator
3.Answer Generator
@\colorbox{orange}{Information gathered by \textbf{Map Service} module via Map-Tool Agent (invoking Trip tool):}@
All Location Info: 
Planetshakers Church Melbourne City Campus 
Name: Planetshakers Church Melbourne City Campus
Address: 400 City Rd, Southbank VIC 3006, Australia
Rating: 4.8
Types: church, tourist_attraction, place_of_worship, point_of_interest, establishment
Is Open Now: False
Weekday Opening Hours:
- Monday: Closed
....
....
- Saturday: Closed
- Sunday: 9:00AM - 8:00PM
Carrot Tattoo Studio 
Name: Carrot Tattoo Studio
Address: Basement/63 Elizabeth St, Melbourne VIC 3000, Australia
Rating: 5
Types: store, point_of_interest, establishment
Is Open Now: False
Weekday Opening Hours:
- Monday: 10:00 AM - 8:00 PM
- Tuesday: 10:00AM - 8:00PM
....
....
- Sunday: 10:00AM - 8:00PM
Melbourne Lifestyle Apartments - Docklands Accommodation 
Name: Melbourne Lifestyle Apartments - Best Views on Collins
Address: 893A Collins St, Docklands VIC 3008, Australia
Rating: 4.8
Types: lodging, point_of_interest, establishment
Is Open Now: True
Weekday Opening Hours:
- Monday: Open 24 hours
- Tuesday: Open 24 hours
....
....
The travel time(distance) from Planetshakers Church Melbourne City Campus to Carrot Tattoo Studio is 9 mins (2.0 km)
The travel time(distance) from Planetshakers Church Melbourne City Campus to Melbourne Lifestyle Apartments - Docklands Accommodation is 8 mins (2.5 km)
The travel time(distance) from Carrot Tattoo Studio to Planetshakers Church Melbourne City Campus is 9 mins (1.9 km)
....
....
The travel time(distance) from Melbourne Lifestyle Apartments - Docklands Accommodation to Carrot Tattoo Studio is 11 mins (2.7 km)

@\colorbox{pink}{Solution generated by \textbf{solution\_generator} module:}@
To determine the total travel time from Planetshakers Church Melbourne City Campus to Carrot Tattoo Studio and then to Melbourne Lifestyle Apartments - Docklands Accommodation, we need to add the travel times for each segment of the journey.

1. **From Planetshakers Church Melbourne City Campus to Carrot Tattoo Studio:**
   - Travel time: 9 minutes

2. **From Carrot Tattoo Studio to Melbourne Lifestyle Apartments - Docklands Accommodation:**
   - Travel time: 10 minutes

Now, we add these travel times together:
9 minutes + 10  minutes = 19 minutes

Therefore, the total travel time is 19 minutes.

@\colorbox{blu}{Correct option extracted by \textbf{answer\_generator} module:}@  @\colorbox{green}{B.19 minutes}@
\end{lstlisting}
\end{figure*}

\begin{figure*}[t]
\begin{lstlisting}[caption={Example how MapAgent answer {\toolnospace-Visual} counting query \colorbox{green}{Green}: Correct Answer. \colorbox{red}{Red}: Wrong Answer.},captionpos=t, basicstyle=\scriptsize\ttfamily, escapeinside={@}{@}, literate={hrule}{{\hrulefill}}5, label=lst:visual-exm-1, breaklines=true, breakindent=0pt, xleftmargin=0pt, frame=lines]
@\colorbox{lightblue}{Context:}@
@\includegraphics[width=1\linewidth]{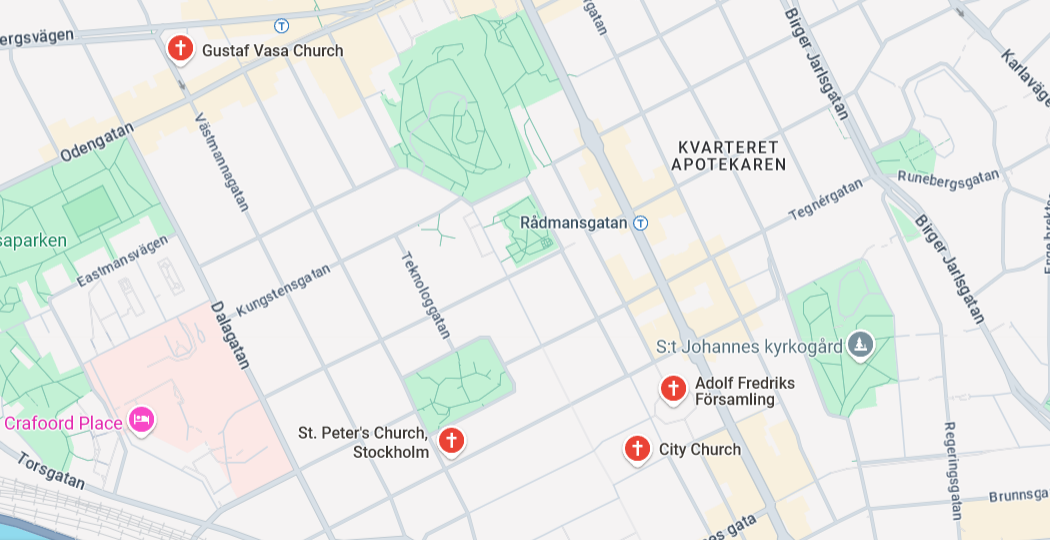}@
@\colorbox{lightblue}{Question:}@ 
How many churces are visible in the lower half of the map?
@\colorbox{yellow}{Options:}@ A.Five B.Two C.Four @\colorbox{green}{D.Three}@

hrule
@\colorbox{lightgray}{Modules selected by \textbf{Planner agent}:}@  
Module Selection from the Module Catalog and their execution order:
1.Visual Place Recognizer module
2.map_service module
3.solution_generator module
4.answer_generator module
hrule
@\colorbox{giallo}{Visual Place Recognizer module Output:}@
This image center represents the location of Radmansgatan 46, 113 57 Stockholm, Sweden.

@\colorbox{orange}{Map Service module Output:}@
When executing the map_service module, it connects to the map-tool agent in the Inner Layer.

@\colorbox{orange}{Map-Tool Agent Output:}@
Nearby tool is selected to fetch information.

@\colorbox{orange}{Nearby Tool's Output:}@
There are some churches within 0.8 kilometers of the location Radmansgatan 46, 113 57 Stockholm, Sweden, as shown below:
City Church ( distance: 0.46526450497253263 kilometers, rating:4.6, total reviewer:94)
St. Peter's Church, Stockholm ( distance: 0.584886996790922 kilometers, rating:4.5, total reviewer:42)
Adolf Fredriks Forsamling ( distance: 0.36249979086127565 kilometers, rating:4.5, total reviewer:341)
hrule

@\colorbox{pink}{solution generator module's Output:}@
Based on the provided map and Google Maps context, the churches visible in the lower half are:
St. Peter's Church, Stockholm
Adolf Fredriks Forsamling
City Church 

Thus, the correct answer is B. Three.

hrule

@\colorbox{blu}{answer generator module's Output:}@
Based on the response from the solution generator, the answer should be 3, which seems logical and correct. The answer is @\colorbox{green}{B. Three}@. 

hrule
@\colorbox{gray}{Comments:}@
The Final answer B. 3 is a correct Answer
\end{lstlisting}
\end{figure*}

\begin{figure*}[t]
\begin{lstlisting}[caption={Example how MapAgent answer {\toolnospace-Visual} Route query \colorbox{green}{Green}: Correct Answer. \colorbox{red}{Red}: Wrong Answer.},captionpos=t, basicstyle=\scriptsize\ttfamily, escapeinside={@}{@}, literate={hrule}{{\hrulefill}}5, label=lst:visual-exm-2, breaklines=true, breakindent=0pt, xleftmargin=0pt, frame=lines]
@\colorbox{lightblue}{Context:}@
@\includegraphics[width=1\linewidth]{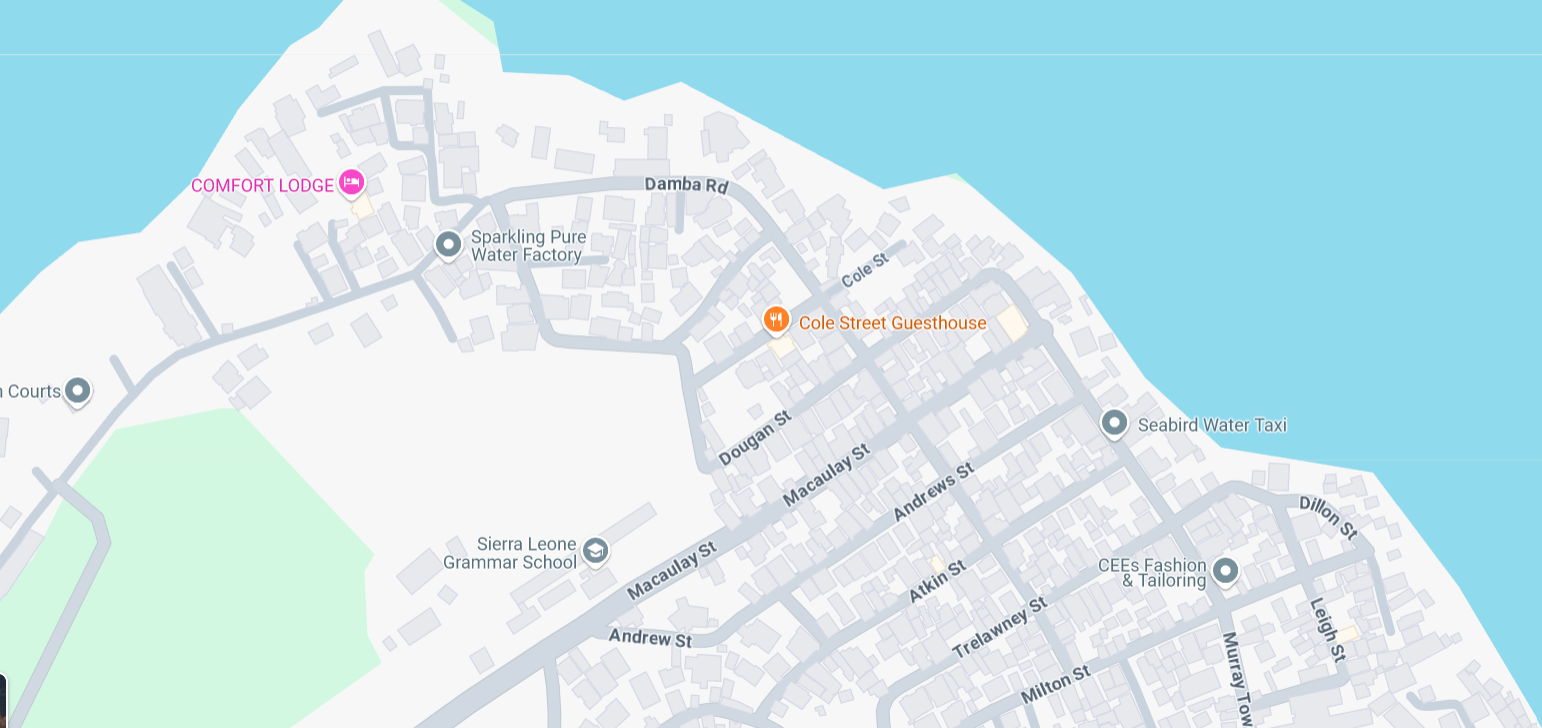}@
@\colorbox{lightblue}{Question:}@ 
How to get to Sierra Leone Grammer School from Seabird Water Taxi after having lunch at Cole Street Guesthouse?
@\colorbox{yellow}{Options:}@
A. Seabird Water Taxi -> Head northwest on High Broad St toward Andrews St -> Turn left onto Andrews St -> Continue onto Andrew St -> Turn right onto Macaulay St -> Sierra Leone Grammar School -> Head northeast on Macaulay St -> Turn left at the 1st cross street toward Cole St -> Turn left at the 2nd cross street onto Cole St -> Cole Street Guesthouse

B. Seabird Water Taxi -> Head northwest on High Broad St toward Andrews St -> Turn left onto Andrews St -> Continue onto Andrew St -> Turn right onto Macaulay St -> Sierra Leone Grammar School

C. Seabird Water Taxi -> Head northwest on High Broad St toward Andrews St -> Turn left onto Macaulay St -> Sierra Leone Grammar School -> Head northeast on Macaulay St -> Turn left at the 1st cross street toward Cole St -> Turn left at the 2nd cross street onto Cole St -> Cole Street Guesthouse

@\colorbox{green}{D.}@ Seabird Water Taxi -> Head northwest on High Broad St toward Andrews St -> Turn right toward Cole St -> Turn left at the 1st cross street onto Cole St -> Cole Street Guesthouse -> Head northeast on Cole St toward Damba Rd -> Turn right at the 1st cross street toward Macaulay St -> Turn right at the 2nd cross street onto Macaulay St -> Sierra Leone Grammar School

hrule
@\colorbox{lightgray}{Modules selected by \textbf{Planner agent}:}@ 
Module Selection from the Module Catalog and their execution order:
1.Visual Place Recognizer module
2.map_service module
3.solution_generator module
4.answer_generator module
hrule
@\colorbox{giallo}{Image Search module Output:}@
This image center represents the location of Cole Street Guesthouse, 5 Cole St, Freetown, Sierra Leone.

@\colorbox{orange}{Map Service module Output:}@
When executing the map_service module, it connects to the map-tool agent in the Inner Layer.

@\colorbox{orange}{Map-Tool Agent Output:}@
Route tool is selected to fetch information.

@\colorbox{orange}{Route Tool's Output:}@
There is a total of one route from Seabird Water Taxi, Freetown, Sierra Leone, to Sierra Leone Grammar School, Freetown, Sierra Leone, with a waypoint at Cole Street Guesthouse. The route information is provided below:
Route 1:(VIA) High Broad St (0.4 km, 5 mins)
Details steps are provided below: 
Head northwest on High Broad St toward Andrews St (0.3 km, 4 mins) 
Turn right toward Cole St (56 m, 1 min) 
Turn left onto Cole StDestination will be on the left (46 m, 1 min) 
Head northeast on cole st toward Damba Rd (46m, 1 min)
Turn right at the 1st cross street toward Macaulay St  (110 m, 1.5 min)
Turn right at the 2nd cross street onto Macaulay St (260 m, 2 min)
Sierra Leone Grammar School
\end{lstlisting}
\end{figure*}

\begin{figure*}[t]
\begin{lstlisting}[caption={Continuation of Example how MapAgent answer {\toolnospace-Visual} Route query \colorbox{green}{Green}: Correct Answer. \colorbox{red}{Red}: Wrong Answer.},captionpos=t, basicstyle=\scriptsize\ttfamily, escapeinside={@}{@}, literate={hrule}{{\hrulefill}}5, label=lst:visual-exm-2-cntd, breaklines=true, breakindent=0pt, xleftmargin=0pt, frame=lines]
@\colorbox{pink}{solution generator module's Output:}@
Based on the provided map and Google Maps context, the best way point is:
Head northwest on High Broad St toward Andrews St 
Turn right toward Cole St 
Turn left onto Cole StDestination will be on the left 
Head northeast on cole st toward Damba Rd 
Turn right at the 1st cross street toward Macaulay St 
Turn right at the 2nd cross street onto Macaulay St 
Sierra Leone Grammar School

Therefor the correct answer is:
**(D) Seabird Water Taxi  
Head northwest on High Broad St toward Andrews St  
Turn right toward Cole St  
Turn left at the 1st cross street onto Cole St  
Cole Street Guesthouse  
Head northeast on Cole St toward Damba Rd  
Turn right at the 1st cross street toward Macaulay St  
Turn right at the 2nd cross street onto Macaulay St  
Sierra Leone Grammar School**

hrule

@\colorbox{blu}{answer generator module's Output:}@
Based on the response from the solution generator, the answer should be D, which seems logical and correct. The answer is @\colorbox{green}{D}@. Seabird Water Taxi -> Head northwest on High Broad St toward Andrews St -> Turn right toward Cole St -> Turn left at the 1st cross street onto Cole St -> Cole Street Guesthouse -> Head northeast on Cole St toward Damba Rd -> Turn right at the 1st cross street toward Macaulay St -> Turn right at the 2nd cross street onto Macaulay St -> Sierra Leone Grammar School

hrule
@\colorbox{gray}{Comments:}@
The Final answer D is a correct Answer
\end{lstlisting}
\end{figure*}

\begin{figure*}[t]
\begin{lstlisting}[caption={Example how MapAgent answer {\toolnospace-Visual} Nearby query \colorbox{green}{Green}: Correct Answer. \colorbox{red}{Red}: Wrong Answer.},captionpos=t, basicstyle=\scriptsize\ttfamily, escapeinside={@}{@}, literate={hrule}{{\hrulefill}}5, label=lst:visual-exm-3, breaklines=true, breakindent=0pt, xleftmargin=0pt, frame=lines]
@\colorbox{lightblue}{Context:}@
@\includegraphics[width=1\linewidth]{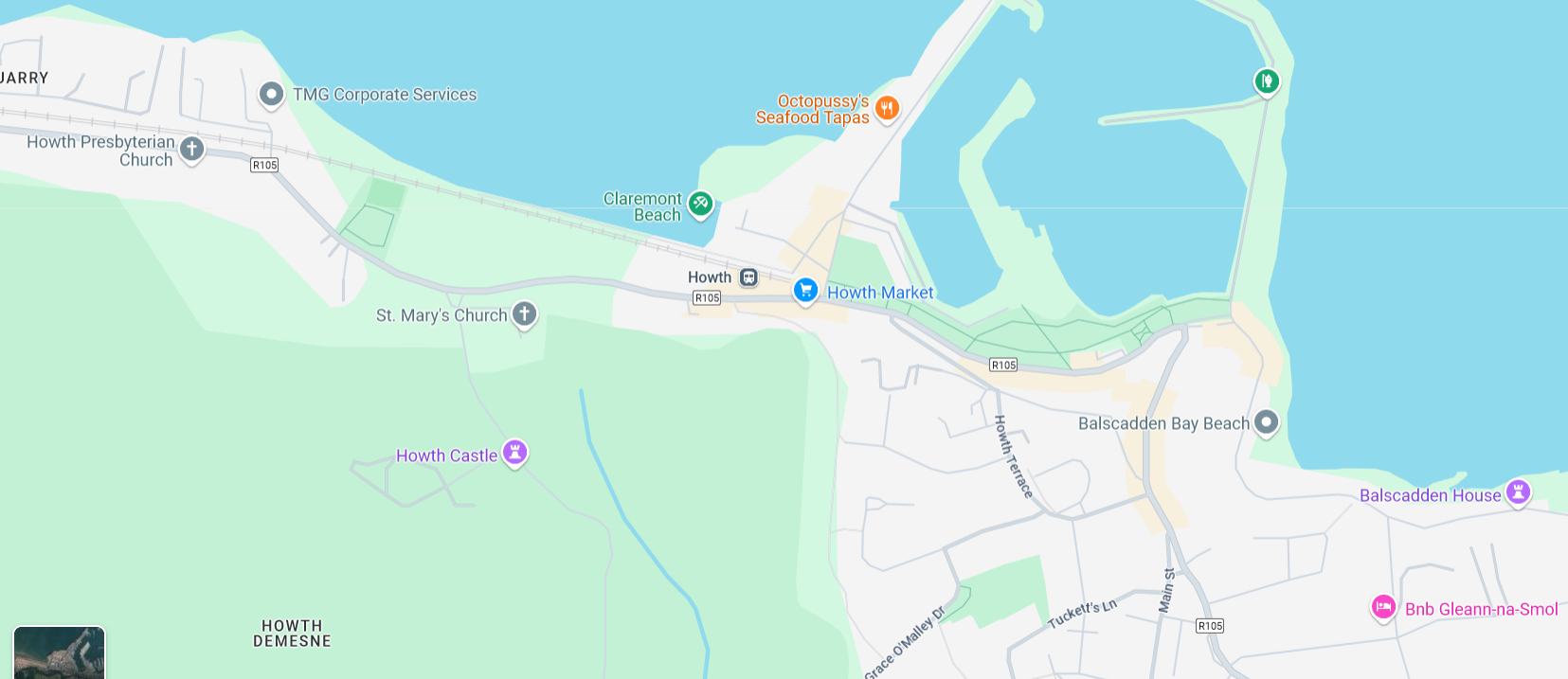}@
@\colorbox{lightblue}{Question:}@ 
Which castle is closest to Howth rail station?
@\colorbox{yellow}{Options:}@  @\colorbox{green}{A.Howth Castle}@  B.Mary Castle  C.Claremont Castle  D.Balscadden House 

hrule
@\colorbox{lightgray}{Modules selected by \textbf{Planner agent}:}@  
Module Selection from the Module Catalog and their execution order:
1.visual_place_recognizer module
2.map_service module
3.solution_generator module
4.answer_generator module
hrule
@\colorbox{giallo}{Visual Place Recognizer module Output:}@
The center of the image appears to be Howth, Ireland.
@\colorbox{orange}{Map Service module Output:}@
When executing the map_service module, it connects to the map-tool agent in the Inner Layer.
@\colorbox{orange}{Map-Tool Agent Output:}@
Nearby tool is selected to fetch information.
@\colorbox{orange}{Nearby Tool's Output:}@
There are some castle distance from the current location Howth Market, 3 Harbour Rd, Howth, Dublin, D13 T8Y7, Ireland within 15 Kilometers in below:
Howth Castle Estate ( distance: 0.4770576879845442 kilometers, rating:4.1, total reviewer:714)
Malahide Castle & Gardens ( distance: 8.742973877916125 kilometers, rating:4.6, total reviewer:15235)
Dublin Castle ( distance: 13.875697894024961 kilometers, rating:4.3, total reviewer:33876)
Swords Castle ( distance: 12.548387008403804 kilometers, rating:4.5, total reviewer:1764)
Dalkey Castle & Heritage Centre ( distance: 12.505523354933828 kilometers, rating:4.6, total reviewer:911)
Monkstown Castle ( distance: 12.235452667377206 kilometers, rating:4.4, total reviewer:104)
Clontarf Castle Hotel ( distance: 9.26724243435518 kilometers, rating:4.5, total reviewer:5073)
Sandycove Castle ( distance: 11.6217343658077 kilometers, rating:4.5, total reviewer:24)
Castle ( distance: 6.943123269967969 kilometers, rating:4, total reviewer:2)
Wooden Castle ( distance: 8.662582766435369 kilometers, rating:4.6, total reviewer:11)
Lambay Castle ( distance: 11.901973302290772 kilometers, rating:4.3, total reviewer:27)
Sutton Castle ( distance: 2.3206824179509207 kilometers, rating:4.2, total reviewer:9)
Archbold's Castle ( distance: 12.507363732093697 kilometers, rating:4.3, total reviewer:8)
hrule

@\colorbox{pink}{solution generator module's Output:}@
Based on the provided map and Google Maps context, The castle closest to Howth rail station is  Howth Castle which is only 0.477 km away from Howth rail station.

Thus, the correct answer is A. Howth Castle

hrule

@\colorbox{blu}{answer generator module's Output:}@
Based on the response from the solution generator, the answer should be Howth Castle, which seems logical and correct. The answer is @\colorbox{green}{A. Howth Castle}@. 

hrule
@\colorbox{gray}{Comments:}@
The Final answer A. Howth Castle is a correct Answer
\end{lstlisting}
\end{figure*}

\begin{figure*}[t]
\begin{lstlisting}[caption={Example 1 how VLLM used API to guess correct answer \colorbox{green}{Green}: Correct Answer. \colorbox{red}{Red}: Wrong Answer.},captionpos=t, basicstyle=\scriptsize\ttfamily, escapeinside={@}{@}, literate={hrule}{{\hrulefill}}5, label=lst:visual1, breaklines=true, breakindent=0pt, xleftmargin=0pt, frame=lines]
@\colorbox{lightblue}{Context:}@
@\includegraphics[width=1\linewidth]{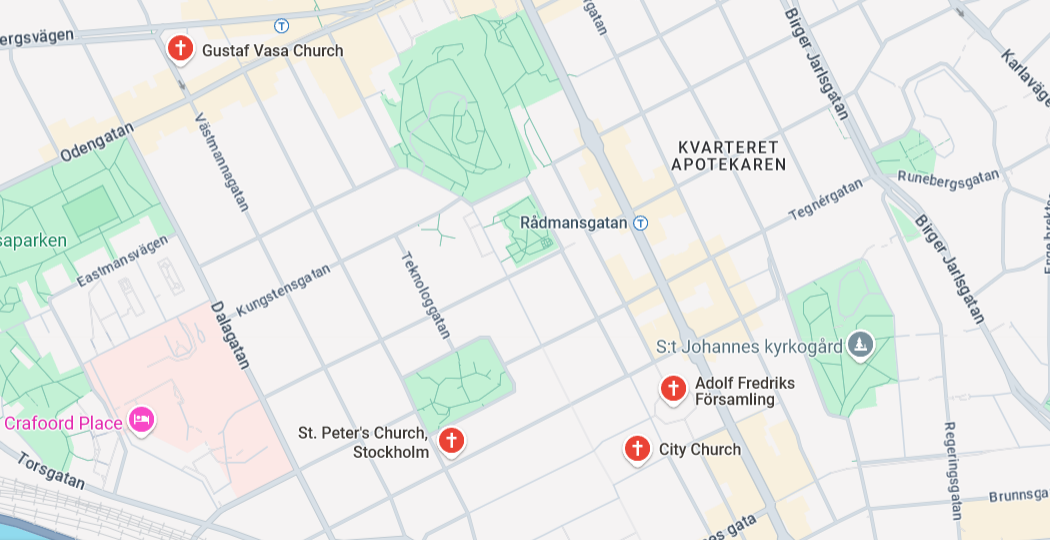}@
@\colorbox{lightblue}{Question:}@ 
How many churces are visible in the lower half of the map?
@\colorbox{yellow}{Options:}@ A.Five B.Two C.Four @\colorbox{green}{D.Three}@

hrule
@\colorbox{lightgray}{Response with using API(MapAgent):}@ 

@\colorbox{pink}{Google Map Response:}@

There are some churches within 0.8 kilometers of the location Radmansgatan 46, 113 57 Stockholm, Sweden, as shown below:
City Church ( distance: 0.46526450497253263 kilometers, rating:4.6, total reviewer:94)
St. Peter's Church, Stockholm ( distance: 0.584886996790922 kilometers, rating:4.5, total reviewer:42)
Adolf Fredriks Forsamling ( distance: 0.36249979086127565 kilometers, rating:4.5, total reviewer:341)
hrule
@\colorbox{blu}{final answer from solution generator:}@

"Based on the map provided and the context from the google map module, the churches visible in the lower half of the map are:"
"1. St. Peter's Church, Stockholm "
"2. Adolf Fredriks Frsamling "
"3. City Church  "
"Thus, the correct answer is:  @\colorbox{green}{(D) Three**}@"


@\colorbox{giallo}{Response without using API(Base Model):}@

Based on the provided map, there are four visible churches in the map. Thus, the answer is: @\colorbox{red}{(C) Four}.@


@\colorbox{gray}{Observation:}@
The correct answer should be D. Three which api version can predict but the base model which do not have api not able to answer.
\end{lstlisting}
\end{figure*}

\begin{figure*}[t]
\begin{lstlisting}[caption={Example 2 how VLLM used API to guess correct answer \colorbox{green}{Green}: Correct Answer. \colorbox{red}{Red}: Wrong Answer.},captionpos=t, basicstyle=\scriptsize\ttfamily, escapeinside={@}{@}, literate={hrule}{{\hrulefill}}5, label=lst:visual2, breaklines=true, breakindent=0pt, xleftmargin=0pt, frame=lines]
@\colorbox{lightblue}{Context:}@
@\includegraphics[width=1\linewidth]{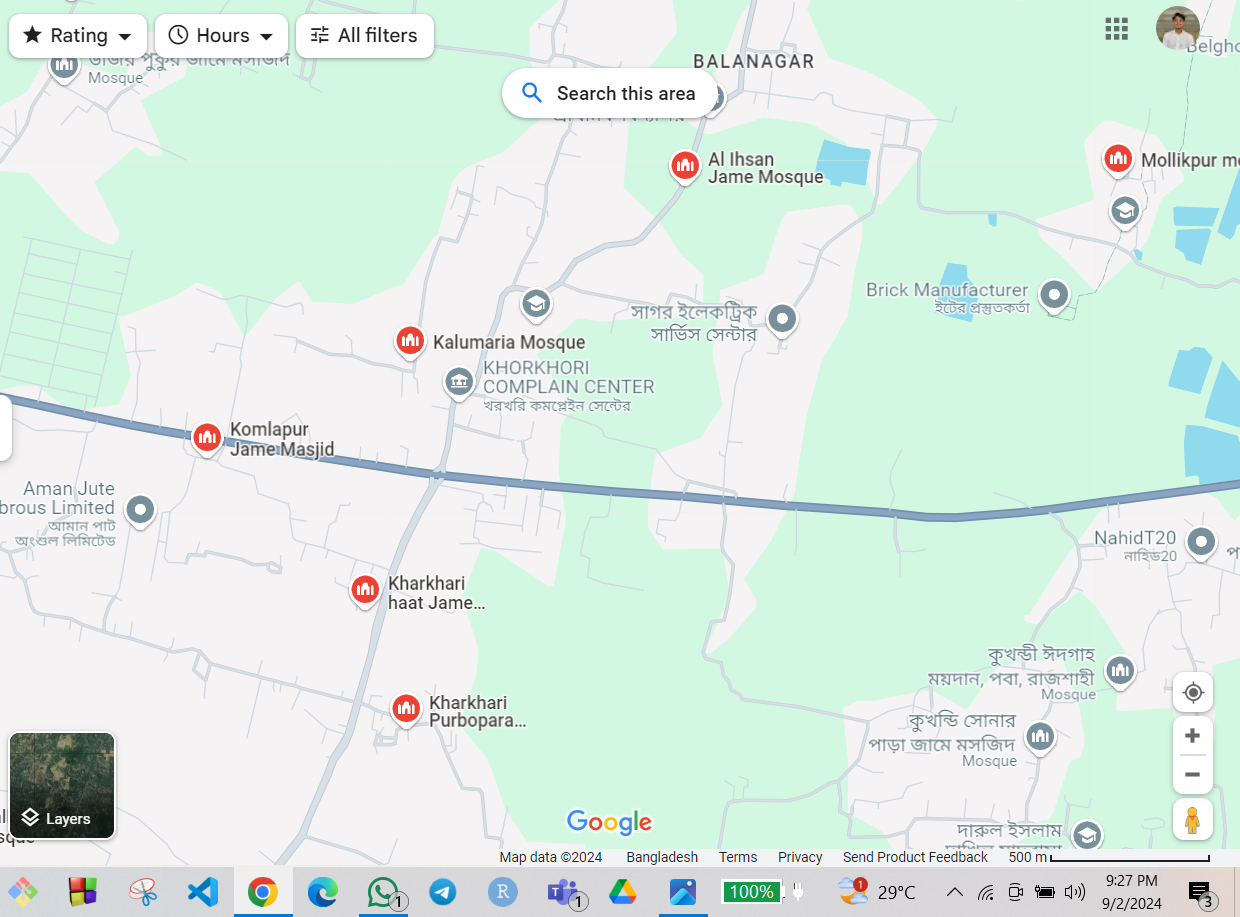}@
@\colorbox{lightblue}{Question:}@ 
I am at Sagor Electronix service center. What's the nearest mosque to me?
@\colorbox{yellow}{Options:}@ A.Kalumaria Mosque @\colorbox{green}{B.Al Ihsan Jame Mosque}@ C.Komlapur Jame Masjid D.Mollikpur mosque
hrule
@\colorbox{lightgray}{Response with using API(MapAgent):}@ 

@\colorbox{pink}{Google Map Response:}@
There are some mosque distance from the current location  Sagor Electronix service center in below:
Al Ihsan Jame Mosque ( distance: 0.5823095053737362 kilometers, rating:5, total reviewer:1)
Mollikpur mosque ( distance: 1.1680710806133061 kilometers, rating:4.7, total reviewer:3)
Kalumaria Mosque ( distance: 1.1879085293895582 kilometers, rating:4.3, total reviewer:4)
Lolithar Bazar Central Masjid ( distance: 1.1566211819806247 kilometers, rating:0, total reviewer:0)
Komlapur Jame Masjid ( distance: 1.8654544610349613 kilometers, rating:4.4, total reviewer:25)
Rajshahi University Central Mosque ( distance: 4.459173025195116 kilometers, rating:4.9, total reviewer:221)
Maria Jama Masjid ( distance: 1.8367913382567742 kilometers, rating:4, total reviewer:9)
Maherchandi East Jame Mosque ( distance: 2.3601717780592995 kilometers, rating:5, total reviewer:5)
Vodra Central Jame Mosque ( distance: 5.131809940553015 kilometers, rating:4.6, total reviewer:145)
Madina Nagar Jame Mosque ( distance: 4.21188964565389 kilometers, rating:4.4, total reviewer:57)
Purbo Para Jame Mosque ( distance: 3.614167437327442 kilometers, rating:4.2, total reviewer:26)
Meherchandi Purbapara Central Jame Mosque ( distance: 3.4298745224188134 kilometers, rating:4.5, total reviewer:69)
Katakhali Bazar Jame Masjid ( distance: 4.28092651862087 kilometers, rating:4.5, total reviewer:193)
Belghoria Jama Masjid ( distance: 5.697832533309013 kilometers, rating:4.6, total reviewer:19)
Hazrapukur Jame Mosque ( distance: 3.260713598861443 kilometers, rating:4.6, total reviewer:30)
hrule
@\colorbox{blu}{final answer from solution generator:}@
The nearest mosque to your location, is **Al Ihsan Jame Mosque** with a distance of 0.582 kilometers.\n\nAnswer: @\colorbox{green}{**(B) Al Ihsan Jame Mosque**}@

@\colorbox{giallo}{Response without using API(Base Model):}@

Okay, I see your location on the map! The nearest mosque to  Sagor Electronix service center appears to be @\colorbox{red}{(C) Kalumaria Mosque.}@

@\colorbox{gray}{Observation:}@
The correct answer should be B. Three which api version can predict but the base model which do not have api not able to answer.
\end{lstlisting}
\end{figure*}

\end{document}